\documentclass[10pt,twocolumn,letterpaper]{article}

\usepackage[pagenumbers]{cvpr} 

\usepackage{booktabs}
\usepackage{subcaption}
\usepackage{multirow}
\usepackage[export]{adjustbox}

\usepackage[dvipsnames]{xcolor}

\usepackage{pgfplots}
\usepackage{tikz}
\usepackage{array}
\usepackage{booktabs}
\usepackage{xcolor}

\usepackage{lipsum}
\usepackage{calc}

\usepackage{makecell, hhline}

\newcommand{\thickhline}{%
    \noalign {\ifnum 0=`}\fi \hrule height 1pt
    \futurelet \reserved@a \@xhline
}
\newcolumntype{"}{@{\hskip\tabcolsep\vrule width 1pt\hskip\tabcolsep}}
\makeatother

\usepackage{array}
\newcolumntype{x}[1]{>{\centering\arraybackslash\hspace{0pt}}p{#1}}

\usepackage{listings}

\usepackage{lineno}
\definecolor{codegreen}{rgb}{0,0.6,0}
\definecolor{codegray}{rgb}{0.5,0.5,0.5}
\definecolor{codepurple}{rgb}{0.58,0,0.82}
\definecolor{backcolour}{rgb}{0.95,0.95,0.92}

\lstdefinestyle{mystyle}{
    backgroundcolor=\color{backcolour},   
    commentstyle=\color{codegreen},
    keywordstyle=\color{magenta},
    numberstyle=\tiny\color{codegray},
    stringstyle=\color{codepurple},
    basicstyle=\ttfamily\footnotesize,
    breakatwhitespace=false,         
    breaklines=true,                 
    captionpos=b,                    
    keepspaces=true,                 
    numbers=left,                    
    numbersep=5pt,                  
    showspaces=false,                
    showstringspaces=false,
    showtabs=false,                  
    tabsize=2
}

\DeclareMathOperator*{\Warp}{\mathcal{W}}
\DeclareMathOperator*{\Linearize}{\mathcal{D}^{-1}}

\DeclareMathOperator*{\Bayer}{\mathcal{M}}
\DeclareMathOperator*{\Loss}{\mathcal{L}}
\DeclareMathOperator*{\RealSet}{\mathbb{R}}

\definecolor{cvprblue}{rgb}{0.21,0.49,0.74}
\usepackage[pagebackref,breaklinks,colorlinks,allcolors=cvprblue]{hyperref}

\def\paperID{3246} 
\def\confName{CVPR}
\def\confYear{2026}

\title{RAW-Domain Degradation Models for Realistic Smartphone Super-Resolution}

\author{Ali Mosleh \quad
Faraz Ali \quad
Fengjia Zhang \quad
Stavros Tsogkas \\
Junyong Lee \quad
Alex Levinshtein \quad
Michael S. Brown
\\AI Center-Toronto, Samsung Electronics
}

\begin{document}
\maketitle

\begin{abstract}
Digital zoom on smartphones relies on learning-based super-resolution (SR) models that operate on RAW sensor images, but obtaining sensor-specific training data is challenging due to the lack of ground-truth images. Synthetic data generation via ``unprocessing'' pipelines offers a potential solution by simulating the degradations that transform high-resolution (HR) images into their low-resolution (LR) counterparts. However, these pipelines can introduce domain gaps due to incomplete or unrealistic degradation modeling. In this paper, we demonstrate that principled and carefully designed degradation modeling can enhance SR performance in real-world conditions. Instead of relying on generic priors for camera blur and noise, we model device-specific degradations through calibration and unprocess publicly available rendered images into the RAW domain of different smartphones. Using these image pairs, we train a single-image RAW-to-RGB SR model and evaluate it on real data from a held-out device. Our experiments show that accurate degradation modeling leads to noticeable improvements, with our SR model outperforming baselines trained on large pools of arbitrarily chosen degradations. We will make our calibrated kernels and noise models publicly available, to facilitate research on image enhancement for mobile photography.
\end{abstract}

\section{Introduction}
\label{sec:intro}
\begin{figure}[t]
\centering
\includegraphics[angle=0,valign=m,width=0.99\columnwidth, trim={0 4cm 0cm 4cm},clip]{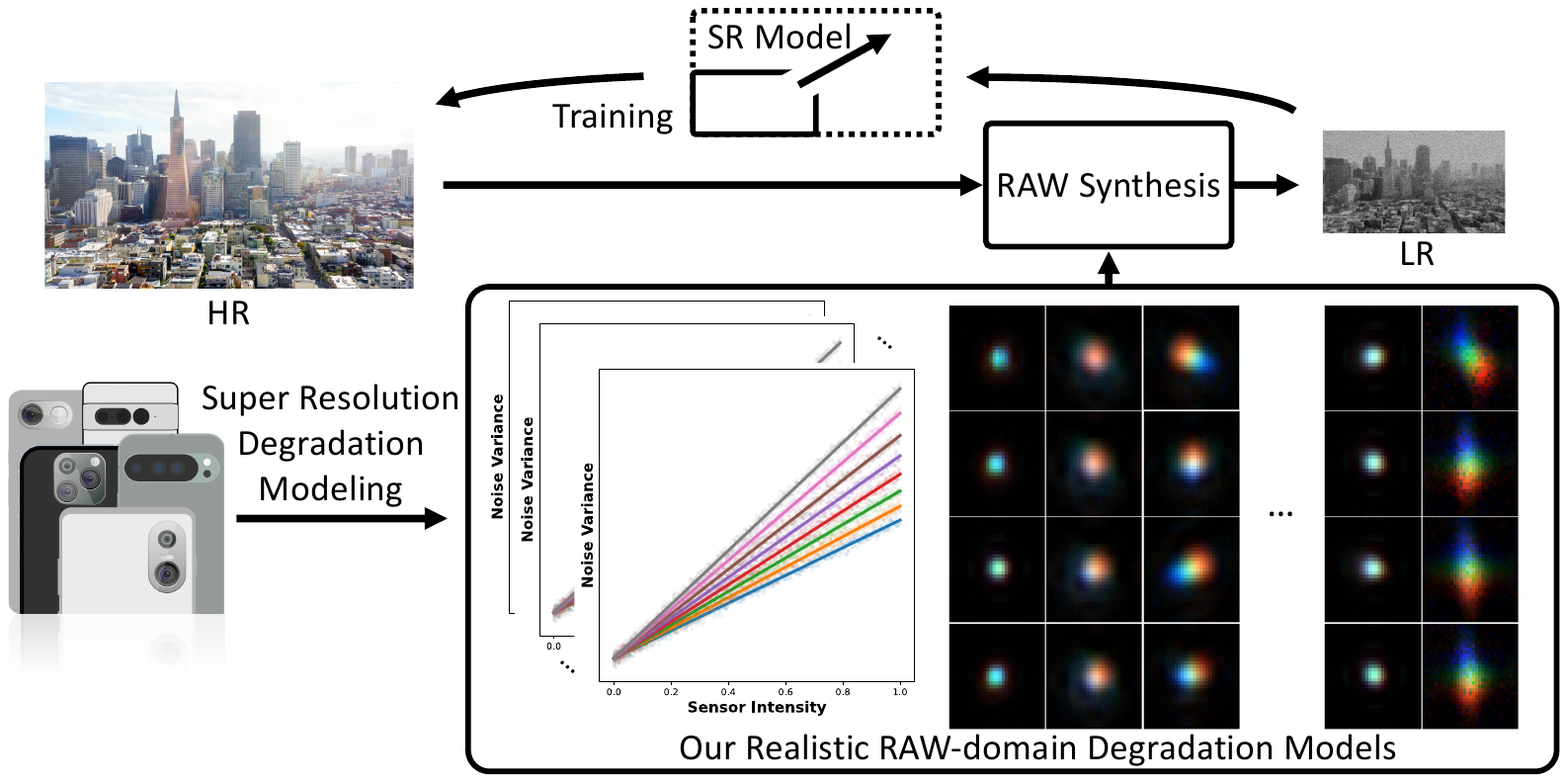} 
\footnotesize
\noindent
\begin{minipage}{0.33\linewidth}
\centering
Original LR Capture 
\end{minipage}%
\begin{minipage}{0.33\linewidth}
\centering
4$\times$ Upsampling 
\end{minipage}%
\begin{minipage}{0.33\linewidth}
\centering
BSRAW~\cite{conde2024bsraw}
\end{minipage}
\centering
\includegraphics[angle=0,valign=m,width=0.99\columnwidth, trim={0 0 0cm 0},clip]{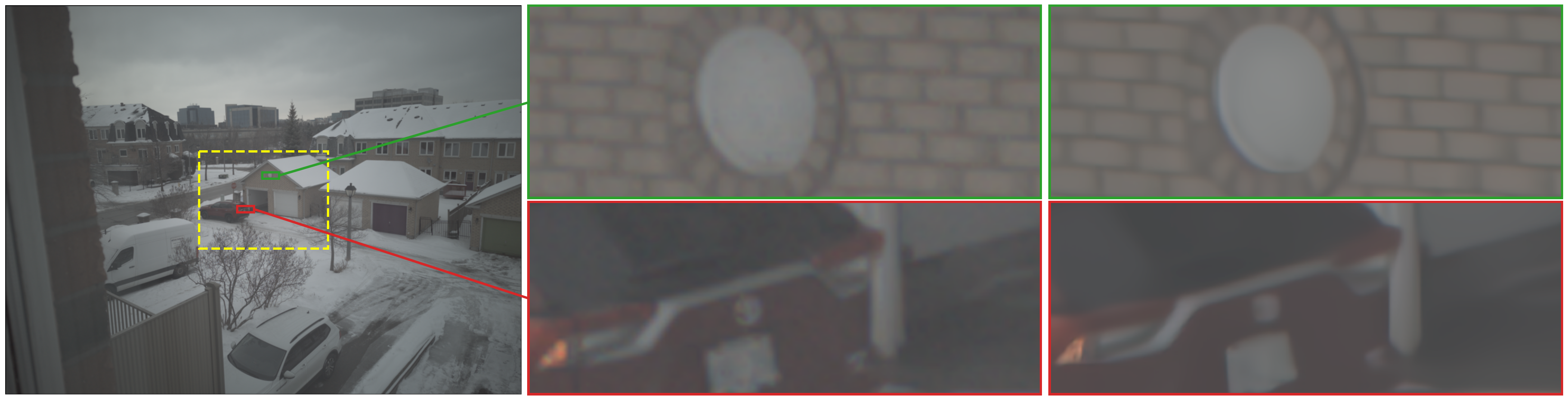}
\begin{minipage}{0.33\linewidth}
\centering
Real-ESRGAN~\cite{wang2021realesrgan} 
\end{minipage}%
\begin{minipage}{0.33\linewidth}
\centering
RAWSR~\cite{xu2019towards} 
\end{minipage}%
\begin{minipage}{0.33\linewidth}
\centering
Ours
\end{minipage}
\includegraphics[angle=0,valign=m,width=0.99\columnwidth, trim={0 0 0cm 0},clip]{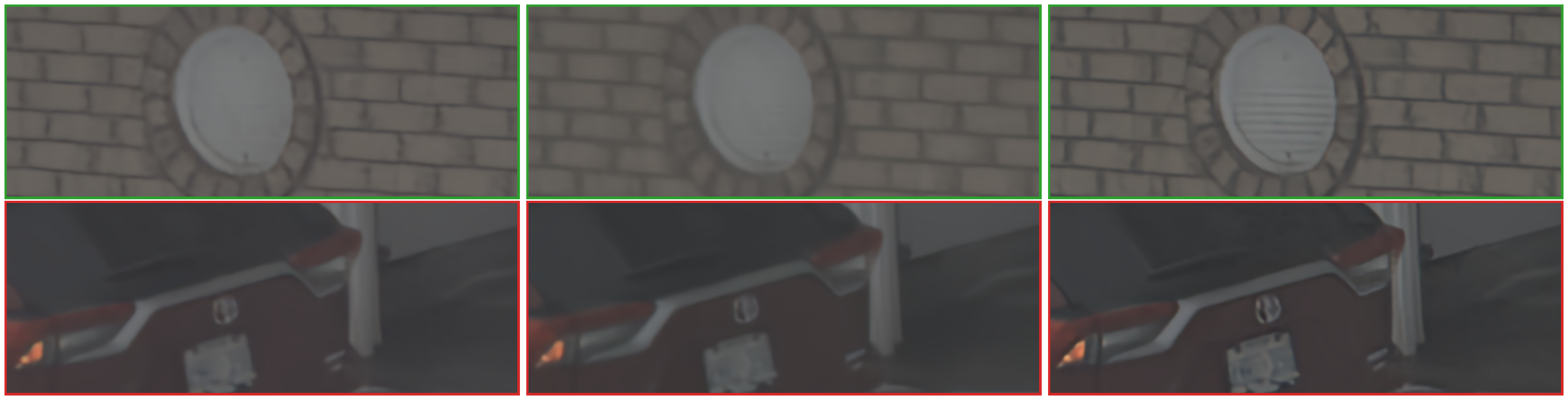}
 \vspace{-0.3cm}
\caption{Our SR approach incorporates realistic SR kernels and sensor noise functions, accurately modeled for various modern 
smartphones. 
A RAW image captured with \textbf{Pixel 6 Main} camera is processed using a 4$\times$ SR model trained 
on data generated with our realistic degradations. 
Our model recovers more details and structures compared to other baselines. 
Pixel 6's specific degradations \emph{are not seen} by the models during training.
\vspace{-0.3cm}}
\label{fig:teaser}
\end{figure}

Modern smartphones have become powerful imaging tools, but their compact design imposes unavoidable optical and sensor limitations.
The limited physical space available for hardware components forces manufacturers to use simpler lens systems and smaller sensors, increasing susceptibility to noise and reducing resolution quality. 
Such limitations are particularly evident in applications such as \emph{digital zoom}, which would benefit significantly from a larger optical system that, however, cannot be accommodated within a compact smartphone form factor.

To compensate for hardware limitations, considerable efforts have been made to replace bulky optical components with software. Image super-resolution (SR) reconstructs a high-resolution (HR) image from a low-resolution (LR) input, effectively mimicking optical zoom. While not a new problem, modern SR methods rely on deep convolutional neural networks (CNNs) trained on large datasets of paired datasets~\cite{dong2015image,anwar2020deep}. However, collecting high-quality data remains challenging.
A common approach captures the same scene with two cameras of varying quality~\cite{ignatov2020replacing} or the same camera at multiple focal lengths~\cite{zhang2019zoom}.
This process is labor-intensive, time-consuming and requires controlled captures with static scenes and precise alignment to avoid artifacts.

A practical alternative is training SR models on synthetically generated data. A common strategy is to downsample HR images from datasets like \cite{Agustsson_2017_CVPR_Workshops,fivek} using a fixed operator (\ie, bicubic interpolation) to create aligned LR-HR pairs. This simplistic degradation model fails to capture the complex optical characteristics of a real imaging system. 
Consequently, SR models trained on bicubic-downsampled data struggle to generalize to real smartphone camera images.

This underscores the importance of accurately modeling the degradation process when synthesizing training data for SR. 
The gap between synthetic and real data mainly arises from two factors: optical blur introduced by the lens system and noise introduced by the camera sensor~\cite{michaeli2013nonparametric,Zhang_2021_ICCV,Yue_2022_CVPR}.
A common approach is to construct a ``pool'' of degradations that are randomly applied to clean HR images to generate their LR counterparts during training~\cite{wang2021realesrgan,Yue_2022_CVPR,conde2024bsraw}.
This strategy enhances the diversity of degradations and improves the generalization of CNN-based models.

However, despite the range of simulated degradations, most existing approaches still fail to capture the true variability present in real imaging devices.
Camera blur is typically modeled using isotropic and anisotropic Gaussian kernels~\cite{xu2019towards,zhang2021designing,conde2024bsraw}, but hyperparameters such as kernel size and standard deviation range can significantly impact SR performance. Moreover, there is no guarantee that these hand-picked parameters accurately represent the point spread functions (PSFs) of a target mobile camera. Noise is generally modeled with either homoscedastic~\cite{zhang2021designing} or heteroscedastic~\cite{xu2019towards}, but both still require appropriate parameter calibration for camera sensors~\cite{SIDD_2018_CVPR}.

In addition, most data synthesis pipelines apply these degradations in the sRGB domain, as sRGB images are easy to obtain and widely used in existing SR frameworks~\cite{wang2021realesrgan}. However, sRGB images have already undergone non-linear and often unknown processing steps, whereas degradations such as blur and noise occur earlier in the camera pipeline, where the relationship between scene radiance and sensor response remains linear. This mismatch introduces a domain gap that reduces SR performance on real images.

Working with RAW data mitigates this issue, as RAW values are directly proportional to scene radiance, allowing for more accurate noise and blur modeling~\cite{zhou2018deep,xu2019towards,bhat2021deep,zhang2019zoom,conde2024bsraw,xu2020exploiting,lecouat2022high,chen2018learning,zhang2021rethinking}. Since most downstream applications require sRGB output and collecting paired RAW-sRGB data is challenging, several studies attempt to invert the camera processing pipeline, ``unprocessing'' publicly available sRGB images into their equivalent RAW versions~\cite{brooks2019unprocessing,seo2023graphics2raw,nam2022learning,afifi2021cie,xing2021invertible,zhang2021designing,conde2024bsraw}. Unprocessing pipelines reverse non-linear operations (\eg gamma correction) and linear operations (\eg white balance, color correction) while also introducing blur, noise, and other degradations.  The more accurately these degradations and ISP processing steps are modeled, the smaller the domain gap between real and synthesized RAW, ultimately improving the performance of SR models trained on de-rendered data.


In this paper, we propose a realistic data synthesis framework for SR, specifically targeting \emph{mobile photography} (\cref{fig:teaser}). Unlike previous works that rely on device-agnostic degradation models, we calibrate blur kernels and noise models for a set of eight mobile cameras. 
We propose a procedure to thoroughly calibrate SR blur kernels and adopt standardized approaches to model sensor noise.
These calibrated degradations are then used to unprocess HR sRGB images into LR RAW data, enabling the training of a single-image RAW-to-sRGB SR model. Our framework is built on the assumption that, while processing pipelines vary across mobile devices, components such as sensors and lenses may share similarities, leading to comparable degradation profiles. 
To validate this hypothesis, we evaluate our model on real RAW images captured by devices whose degradations excluded from the training set. Our results show that the proposed device-specific data synthesis approach more effectively models real-world degradations than methods that use generic blur kernels and noise parameters.

Our contributions are as follows:
\begin{itemize}[left=1.0em]
    \setlength\itemsep{0.01em}
    \item A calibration method for accurately modeling SR blur kernels directly within the mosaicked RAW domain of smartphone cameras. 
    \item A blind SR approach that integrates device-specific degradation models for realistic RAW data synthesis.  
    \item State-of-the-art performance on test images with unknown degradations, demonstrating that degradation profiles overlap across mobile cameras and that accurately modeling these degradations outperforms training with generic degradation pools.  
    \item A comprehensive dataset of calibrated blur kernels and noise models for eight modern mobile cameras.  
\end{itemize}

\section{Related Work}
\label{sec:related_work}

\subsection{RAW Image Super-Resolution}
The vast majority of SR methods use sRGB images as inputs, but there is a growing shift toward leveraging RAW data, as they enable more accurate degradation modeling, leading to better SR results. 
Xu \etal~\cite{xu2019towards} synthesize pairs of HR linear RGB and LR RAW images for training alongside their corresponding color inputs, allowing the decoupling of detail/structure and color restoration. 
Similarly, Zhang \etal~\cite{zhang2019zoom} demonstrate the benefits of RAW data for computational zoom. 
They carefully collect a dataset of RAW image pairs with \emph{optical zoom} ground-truth by varying the focal length of a DSLR zoom lens and use it to train a RAW-to-sRGB SR model. 
Beyond improving SR, using RAW data enables the joint optimization of related tasks, eliminating the need for multiple models while enhancing performance across individual tasks.
A classic example is joint demosaicking and SR~\cite{zhou2018deep}, where HR sRGB images are progressively blurred, downsampled, and mosaicked to generate LR RAW data for training a deep CNN. However, the blur kernels used to model the camera PSF are not well-documented, and sensor noise is omitted from their degradation pipeline.  In contrast, BSRAW~\cite{conde2024bsraw} employs a more controllable degradation pipeline that integrates diverse real-world noise profiles, a broad range of blur kernels (including isotropic/anisotropic Gaussian, real estimated blur kernels, and real estimated PSFs), and simulations of varying exposure levels during capture.

The most closely related works are \cite{xu2019towards} and \cite{conde2024bsraw}, both of which employ comprehensive degradation pipelines. However, they still rely on sampling from a broad pool of synthetic and real degradations. In contrast, we take a more principled approach by using carefully calibrated blur and noise profiles, extracted from various mobile cameras.

\subsection{Degradation Modeling for Super-Resolution}

\paragraph{Implicit degradation modeling} using Generative Adversarial Networks (GANs)~\cite{goodfellow2014generative} has gained popularity, as GAN training does not require paired HR-LR data. Bulat \etal~\cite{bulat2018learn} implicitly model HR-to-LR degradations through a CNN that learns to degrade HR images into their LR counterparts, generating paired data that can then be used to train an SR model in a supervised manner.
CinCGAN~\cite{yuan2018unsupervised} follows a similar approach, learning to map HR images to a ``clean LR'' space that resembles the bicubic-downsampled LR domain used by supervised SR methods. DSGAN~\cite{fritsche2019frequency} first downsamples HR images using bicubic interpolation, then trains a model to translate bicubic-LR into the ``real LR'' domain encountered during testing.

The success of these methods in generating realistic LR images can be attributed to the ability of standard CNNs to approximate various degradation operations. Since most types of image degradation have a corresponding operation within a CNN, the network architecture itself can act as an implicit prior for different degradation types~\cite{luo2023and}.

Specific degradation priors can also be imposed through careful data collection. For instance, Chen \etal~\cite{chen2019camera} model the detail loss caused by the intrinsic resolution across the field of view (FOV) when zooming in and out with an optical lens, and use it as a latent variable in the SR process.
 
\paragraph{Explicit degradation modeling} is widely used in \emph{blind SR}~\cite{liu2022blind}, where the underlying degradation of LR images is unknown. Michaeli and Irani~\cite{michaeli2013nonparametric} recover the optimal blur kernel directly from an LR image by leveraging the recurrence property of small natural image patches. KernelGAN~\cite{bell2019blind} expands on this idea by replacing the fixed downscaling operation with a trainable model that learns the ideal SR downscaling function. Notably, both~\cite{michaeli2013nonparametric, bell2019blind} estimate the optimal blur kernel for a specific input LR image.

An alternative approach is to model degradation distributions~\cite{zhou2019kernel,luo2022learning, zhang2023crafting}. These methods typically use adversarial learning~\cite{goodfellow2014generative}, where a generator predicts realistic blur kernels under the assumption that the blur kernel is spatially invariant for a given image. MANet~\cite{liang2021mutual}, in contrast, estimates spatially varying kernels from small LR patches, enabling a more localized and content-aware solution. Other works make reasonable assumptions about the parametric form of degradation, such as modeling blur kernels as isotropic or anisotropic Gaussians, and then generate a diverse pool of randomly sampled kernels.

Real-ESRGAN~\cite{wang2021realesrgan} is one of the most widely used frameworks of this kind, incorporating synthetic noise and JPEG artifact modeling. Despite its practicality, Real-ESRGAN applies degradations to sRGB images, creating a domain gap between training and real-world testing. In our work, we address this gap by carefully modeling degradations in the RAW domain, which also enables more accurate noise modeling. Various distributions have been proposed in the literature to model camera sensor noise.

Commonly, a Poisson-Gaussian (PG) noise model is considered \cite{foi2008practical,makitalo2012optimal}. Work by \cite{wei2021physics} models shot noise with a Poisson distribution, read noise with a Tukey lambda distribution, row noise with a Gaussian distribution, and quantization noise with a uniform distribution. Work in \cite{zhang2021rethinking} employs a Poisson distribution to model signal-dependent noise, while signal-independent noise is directly sampled from a database of dark frames.  Replacing the Poisson distribution with a Gaussian distribution that has signal-dependent variance results in the heteroscedastic Gaussian (HG) model~\cite{liu2014practical,foi2008practical}, which is more widely used in practice~\cite{brooks2019unprocessing}. We adopt this model to construct our pool of noise models calibrated for various smartphone cameras.

Camera-specific degradation modeling is another approach that involves calibrating intrinsic lens PSFs. This is typically achieved using designated charts to obtain HR targets and their corresponding LR captures in an optimization process~\cite{chen2021extreme,diamond2021dirty,mosleh2015camera,chen2025physics}. Since realistic SR depends on a camera’s inherent optical properties, these methods are highly useful. However, they have two limitations: (1) they are restricted to correcting optical aberrations and are not well-suited for higher resolutions, and (2) they tend to be biased toward the patterns or charts used to obtain image pairs for kernel estimation. To address these limitations, we introduce a technique that models camera-specific super-resolved PSFs across different cameras. This step is essential for establishing our collection of realistic SR kernels.


\section{Method}
\label{sec:method}

The LR image formation can be formulated as 
{
\setlength{\abovedisplayskip}{4pt}
\setlength{\belowdisplayskip}{4pt}
\setlength{\arraycolsep}{1pt}
\medmuskip=0mu
\thinmuskip=0mu
\thickmuskip=0mu
\begin{align}
& \mathbf{y} = 
\Bayer 
\Bigl( 
(\mathbf{K}\mathbf{x} ) 
\downarrow_s 
\Bigl) 
+ 
\mathbf{n},
\label{eq:SR_formulation}
\end{align}
where $\mathbf{x}$ is the latent HR image and $\mathbf{y}$ is the LR image in the sensor RAW domain, downsampled by a factor of $s$. 
$\mathbf{K}$ is the blur kernel representing the HR to LR degradation of the optical system, and $\mathbf{n}$ denotes sensor noise. 
Finally, $\Bayer$  is the mosaicking operator, converting a 3-channel RGB image into a 1-channel mosaicked RAW image, 
according to the color filter array (CFA) of the sensor.}

Given a set of SR kernels and sensor noise models, we follow \cref{eq:SR_formulation} to unprocess HR linear-sRGB images into LR RAW images, forming paired HR-LR data for SR training.
In the next sections, we detail how we model SR kernels $\mathbf{K}$, and sensor noise $\mathbf{n}$, for different mobile cameras.




\begin{figure}
    \centering
    \settototalheight{\dimen0}{ \includegraphics[angle=0,valign=m,width=0.99\columnwidth, trim={0.5cm 2.5cm 0cm 1cm},clip]{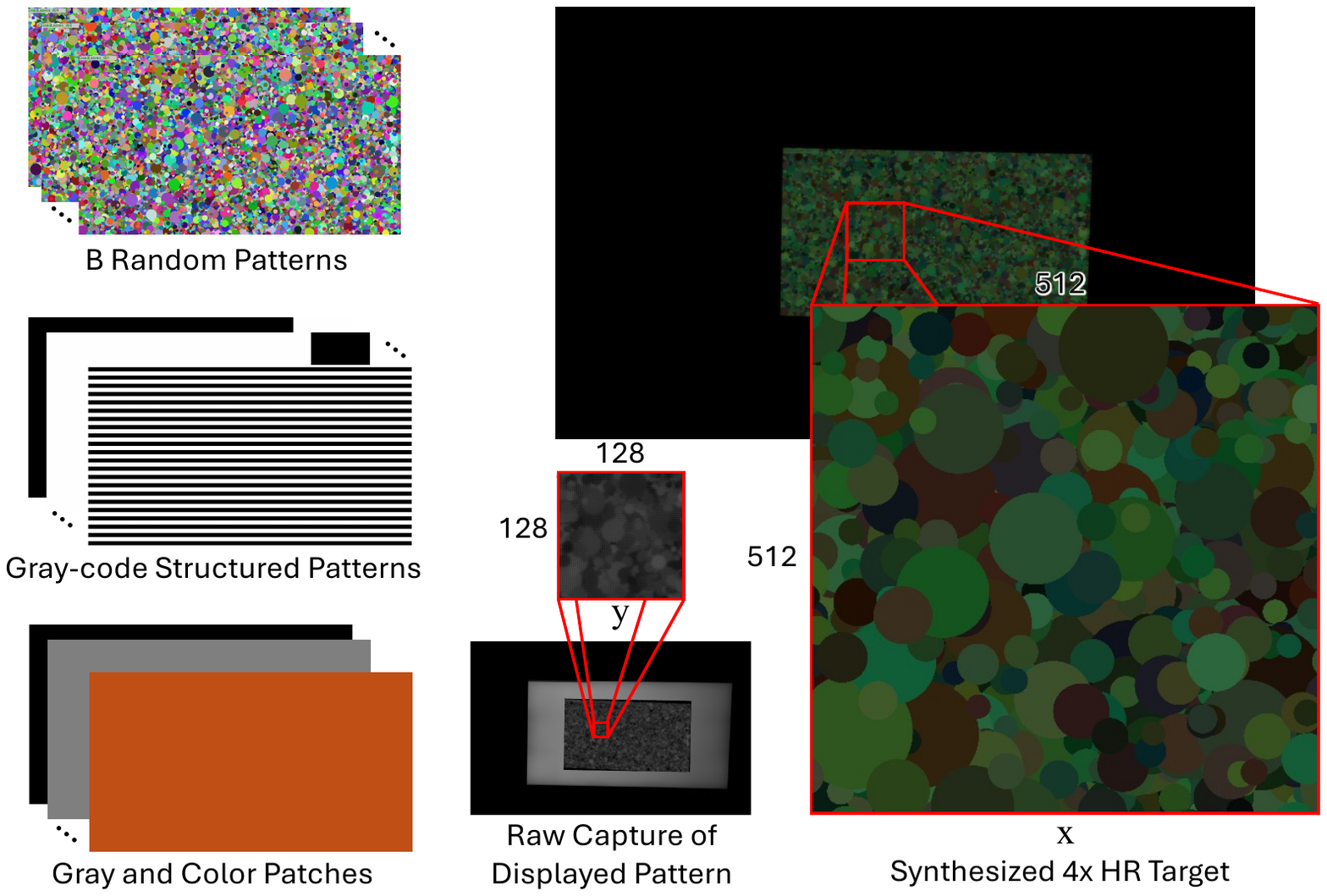} }%
    \includegraphics[width=\columnwidth, trim={0.5cm 2.5cm 0cm 1cm},clip]{figs/fig_calib_patterns.pdf}%
    \hspace*{-0.218cm}\llap{\raisebox{\dimen0-4.98cm}{
      \includegraphics[height=3.1cm]{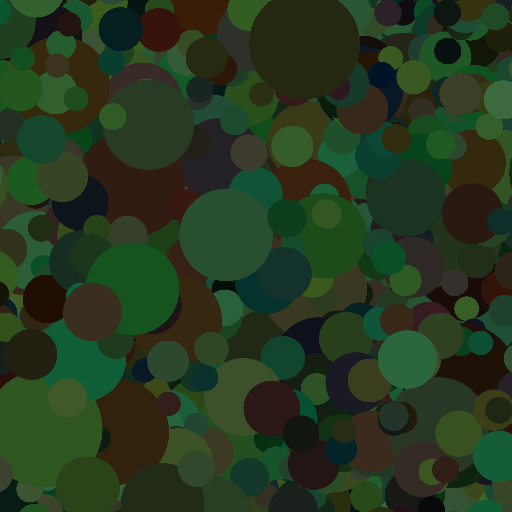}%
    }}
    \vspace{-0.2cm}
\caption{Examples of calibration patterns and images used to approximate SR kernels. Gray-code patterns and color/gray patches displayed on a monitor, along with their RAW capturing with the target camera, are used for geometric and radiometric alignment and to form HR target images. A pair of $\mathbf{x}$ and $\mathbf{y}$ used in \cref{eq:kernel_modeling} for 4$\times$ SR kernel estimation of Pixel 9 Pro Tele is illustrated.\vspace{-0.4cm}
 }
\label{fig:monitor_sensor_image_fromation_examples}
\end{figure}

\subsection{SR Kernel Modeling}
\label{subsec:kernel_modeling}
The SR kernel combines the lens PSF with the discretization operator of the camera sensor~\cite{michaeli2013nonparametric}. As a result, it is influenced not only by the lens characteristics but also by factors such as sensor size and the optical behavior of its micro-lens array. To account for this, we avoid the commonly used SR kernel model—where the lens PSF is followed by bicubic resampling—and instead explicitly model SR kernels using HR-LR paired data collected with a display prototype.

Assuming a noise-free observation, we model imaging from a displayed $M \times M$  HR pattern to an $N \times N$ target camera sensor for SR scale $s$ in vector notation by
{
\setlength{\abovedisplayskip}{4pt}
\setlength{\belowdisplayskip}{4pt}
\setlength{\arraycolsep}{1pt}
\medmuskip=0mu
\thinmuskip=0mu
\thickmuskip=0mu
\begin{align}
& \Tilde{\mathbf{y}} = 
\Bayer 
\Biggl( 
\biggl( 
\begin{bmatrix}
    \mathbf{K}_r       & 0 & 0 \\
          0 & \mathbf{K}_g &  0 \\
    0       & 0 & \mathbf{K}_b
\end{bmatrix}  
\Bigl( \mathbf{v} \odot \Linearize \bigl( \Warp(\acute{\mathbf{x}}, \mathbf{H}) \bigr)  \Bigr) 
\biggr ) \downarrow_s 
\Biggr).
\label{eq:monitor_to_sensor_formation_modeling}
\end{align}
A high-resolution sRGB image $ \acute{\mathbf{x}} \in \RealSet ^{3M^2 \times 1} _{[0, \max_{ \acute{\mathbf{x}}} ]} $  is displayed on a monitor ($M\geq sN$)
and its image is captured by the target camera denoted by $\mathbf{y} \in \RealSet ^{N^2 \times 1}_{[0, 1 ]}$ in black-level/white-level corrected sensor RAW domain, represented in the unit interval. The maximum pixel value in the HR display pattern $\max_{ \acute{\mathbf{x}}}$ is determined by the color depth format used to store it (\eg, $\max_{\acute{\mathbf{x}}} = 2^{16}-1$ in the 16-bit color depth). 
%
%
The geometric projection from $\acute{\mathbf{x}}$ to super-resolved $\mathbf{y}$ is modeled using a perspective transformation 
$\Warp: \RealSet ^{3M^2 \times 1}_{[0, \max_{ \acute{\mathbf{x}}} ]} \times \RealSet ^{3 \times 3} \rightarrow \RealSet ^{3s^2N^2 \times 1}_{[0, \max_{ \acute{\mathbf{x}}} ]}$ as a function of the HR image and a homography matrix $ \mathbf{H} \in \RealSet ^{3 \times 3}$. 
The displayed image undergoes a transformation through the display color space, which is almost always non-linear, before being projected onto the target camera sensor.
This is modeled with a non-linear function $\mathcal{D}(\cdot)$ whose inverse $\Linearize: \RealSet ^{3s^2N^2 \times 1}_{[0, \max_{ \acute{\mathbf{x}}} ]}  \rightarrow \RealSet ^{3s^2N^2 \times 1}_{[0, 1 ]}$
linearizes the warped $\acute{\mathbf{x}}$ and maps it into the sensor's RAW color space.
We also factor in the spatial non-uniformity in brightness projected onto the sensor. This is due to lens vignetting, alongside the varying brightness of the LEDs across the display, 
and is denoted by $ \mathbf{v} \in \RealSet ^{3s^2N^2 \times 1} _{[0, 1 ]} $.}

{
\setlength{\abovedisplayskip}{4pt}
\setlength{\belowdisplayskip}{4pt}
\setlength{\arraycolsep}{1pt}
\medmuskip=0mu
\thinmuskip=0mu
\thickmuskip=0mu
Given $\mathbf{x}=\mathbf{v} \odot \Linearize \bigl( \Warp(\acute{\mathbf{x}}, \mathbf{H}) \bigr) $ as the HR RGB target in \eqref{eq:monitor_to_sensor_formation_modeling}, 
where $\mathbf{x} \in \RealSet ^{3s^2N^2 \times 1}_{[0, 1 ]}$ and $\odot$ denotes element-wise multiplication, we compute kernels for SR scale $s$ as 
\begin{align}
& \{\hat{\mathbf{K}}_r, \hat{\mathbf{K}}_g,\hat{\mathbf{K}}_b\} = 
\underset{ \{\mathbf{K}_r, \mathbf{K}_g,\mathbf{K}_b , \mathbf{H}\}}{\arg\min}
\Loss({\mathbf{y}} , \Tilde{\mathbf{y}} ),
\label{eq:kernel_modeling}
\end{align}
where $ \mathbf{K}_c \in \RealSet ^{s^2N^2 \times s^2N^2}, \forall c \in \{r,g,b\} $ denotes the SR kernel in convolution matrix form for each of the $r$, $g$, and $b$ color channels, and $\Loss(\cdot)$ denotes a regression loss function.}

In practice, the SR kernels vary across the FOV. Therefore, we subdivide it into smaller patches and perform the optimization separately for each one.
Thus, $\mathbf{y}$ corresponds to a small $N$~$\times$~$N$ patch on the sensor. However, a global homography may not accurately align the scene and the sensor due to erroneous planar assumptions, effects of stronger blur kernels on the alignment features in the periphery, \etc. 
Hence, we first use a set of displayed gray-code structured patterns to establish dense correspondences between the sensor pixels and the monitor pixels and calibrate a global homography to initialize $\mathbf{H}$. We then optimize it in \cref{eq:kernel_modeling} per $N \times N$ patch location on the sensor. Our SR kernel modeling is thus a joint optimization problem w.r.t alignment mapping between corresponding patches in the sensor and the display, and color-channel specific SR kernels.

The spatial resolution of the displayed image $\acute{\mathbf{x}}$ must be larger than the target resolution, \ie, $M\geq sN$, to ensure the occurrence of the actual subsampling performed in the sensor, and to avoid any upsampling caused by the warping function. Furthermore, to prevent biasing the kernels towards specific features in the displayed image, we perform optimization on a batch of $B$ HR patterns composed of randomly generated structures and their corresponding LR captures. The ability to use large LR-HR paired data, enabled by careful radiometric and spatial alignment, makes the SR kernel estimation problem well-posed, eliminating the need to explicitly impose heuristic priors or constraints on the kernels, such as non-negativity, energy preservation, sparsity, \etc. This method also facilitates the calculation of loss in the mosaicked RAW domain, preventing domain gap caused by demosaicking and other post-processing steps.

\cref{fig:monitor_sensor_image_fromation_examples} illustrates examples of the random patterns, one of their RAW capture, and its synthesized HR counterpart used in \cref{eq:kernel_modeling} along with a set of displayed patterns for global alignment estimation (initializing $\mathbf{H}$) and gray/color patches for modeling $\Linearize(\cdot)$ and $\mathbf{v} $ across the sensor. 
Refer to the supplemental material for additional details.

\subsection{Sensor Noise Modeling}
\label{subsec:noise_modeling}

\begin{figure}[t]
\centering
\includegraphics[angle=0,valign=m,width=0.32\columnwidth, trim={0.2cm 0.2cm 0.2cm .9cm},clip]{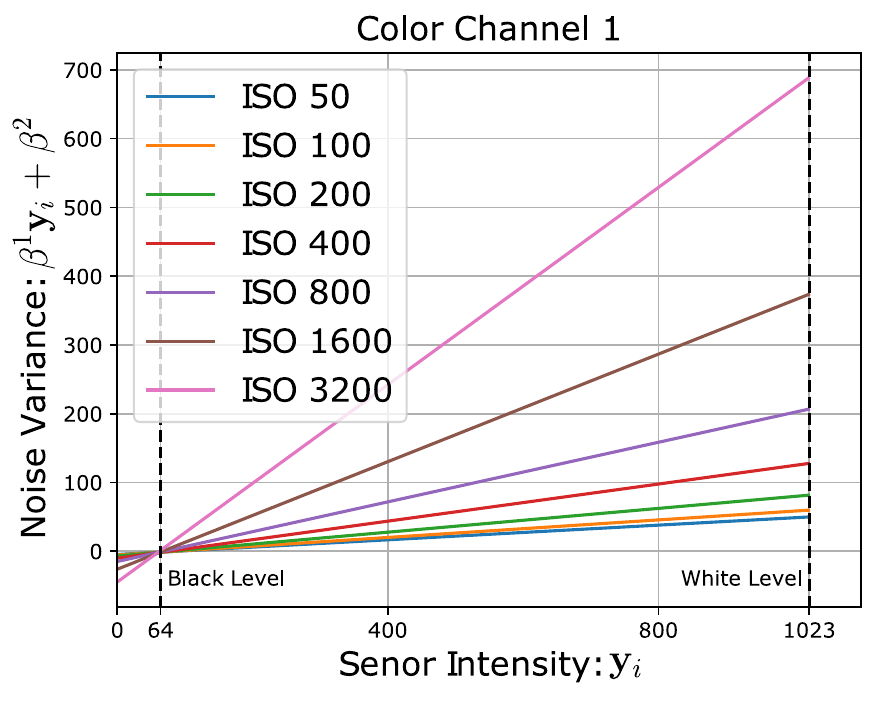}
\includegraphics[angle=0,valign=m,width=0.32\columnwidth, trim={0.2cm 0.2cm 0.2cm 0.9cm},clip]{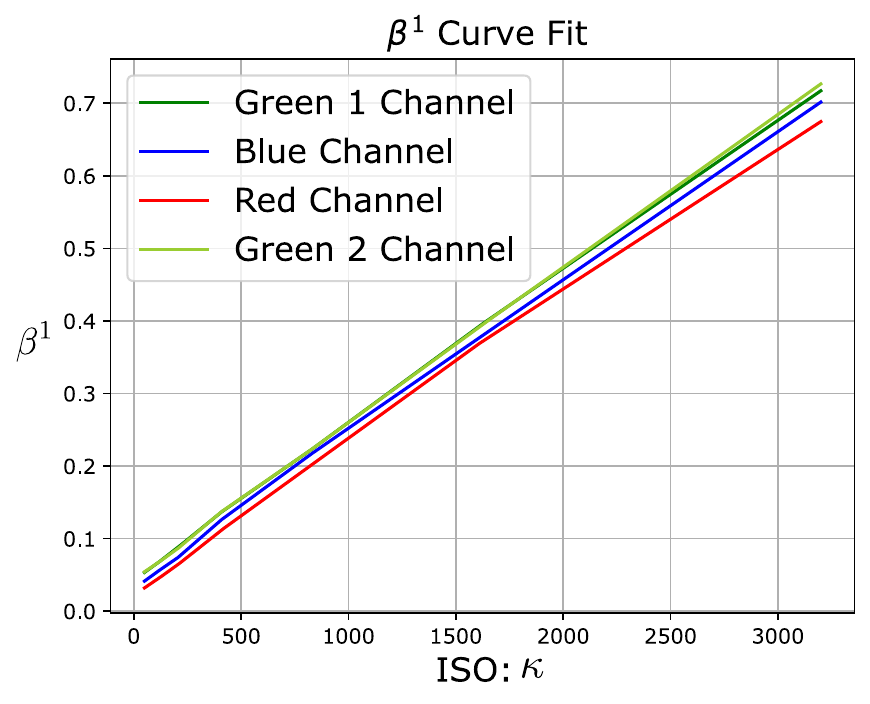}
\includegraphics[angle=0,valign=m,width=0.32\columnwidth, trim={0.2cm 0.2cm 0.2cm 0.9cm},clip]{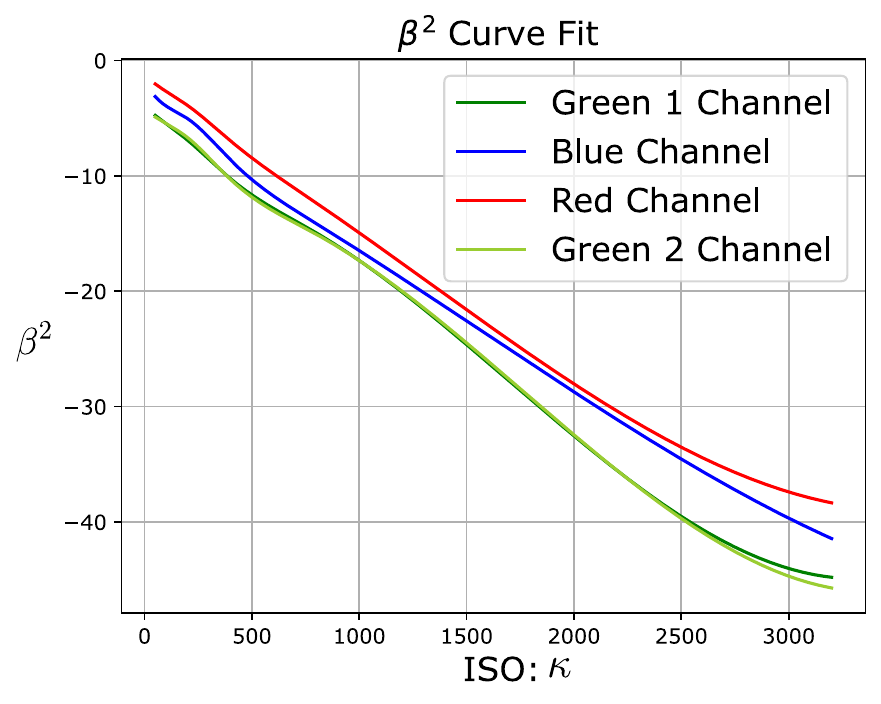}
 \vspace{-0.3cm}
\caption{Noise model calibration results for the S23U Main camera. Noise calibration is performed per color channel of the CFA at seven different ISO levels shown for the first green channel here. A curve is then fit to each noise parameter, enabling interpolation of noise variance for uncalibrated ISO levels.\vspace{-0.3cm}}
\label{fig:noise_model_examples}
\end{figure}

A widely adopted model for noise in modern sensors is a Gaussian distribution with signal-dependent variance~\cite{foi2008practical}. 
This is commonly referred to as the heteroscedastic Gaussian (HG) model and it relates pixel intensity in the noise-free image  $\mathbf{y} = \Bayer \bigl( (\mathbf{K}\mathbf{x} ) \downarrow_s \bigl) $ to shot and read noise parameters $\beta_{\kappa,c}^1$ and $\beta_{\kappa,c}^2$ as $\mathbf{n}_i
\sim
\mathcal{N} \left( 0, \beta_{\kappa,c}^1 \mathbf{y}_i+ \beta_{\kappa,c}^2 \right ).
$
Here $\mathbf{y}_i$ denotes noise-free intensity at pixel index $i$ and  $\mathbf{n}_i$ is a random sample drawn from the normal distribution. 
The noise parameters can be effectively calibrated for each camera. 
However, $\beta_{\kappa,c}^{\{1,2\}}$ vary across different ISO levels and color channels. 
Therefore, we calibrate the model separately per ISO level $\kappa$ and per CFA color channel $c \in \{r,g_1,g_2,b \}$. We model separate noise models for the two green channels in the CFA. Following the recommendation in~\cite{yoshimura2023rawgment}, we run noise sampling along a burst of images for a more accurate calibration. 
One issue is that sensor noise modeling is often performed for a few nominal ISO levels, 
but, in practice, a denser set of ISOs is used to calculate exposure, posing a hindrance for perfect calibration. 
To address this, we fit a quadratic curve to parameters $\beta_{\kappa,c}^{\{1, 2\}}$, calibrated at different ISO levels $\kappa$, and use the fit to approximate noise parameters for uncalibrated ISO values. \cref{fig:noise_model_examples} illustrates the calibrated noise models for the first color channel and the interpolation fits for the noise parameters for one of the cameras we used in experiments.

\subsection{Synthesizing Training Data} 
\label{subssec:method_synthesizing_traing_data}

Our protocol for generating synthetic RAW data from linear RGB images consists of four steps. First, we treat linear RGB images as scene inputs and adjust their pixel values to the target sensor's range, accounting for its black and white levels. Next, we apply a randomly selected RGB blur kernel from our kernel pool and subsample the blurred image at the target SR scale. 
The image is then mosaicked according to the sensor’s CFA pattern. 
To ensure accurate noise modeling, we invert the sensor's white balance gains (obtained during noise calibration)
before adding synthetic noise to each color channel, based on a randomly sampled ISO level.

\section{Experiments}
\label{sec:experiments}
We model degradations for nine different mobile cameras, including Google's Pixel 9 Pro Tele, Pixel 9 Pro Main, Pixel 6 Main, Samsung's S24U Tele 2, S23U Tele 1, S23U Main, S23+ Tele, S23+ Main, and Xiaomi's Mi 11 Main. We also collect evaluation data using these devices. For a detailed list of camera specifications, refer to the supplemental material. To assess the effectiveness of our degradation modeling, we conduct two evaluations. First, we use the modeled degradations to train camera-specific SR models (\cref{subsec:experiments_per_camera}). Second, we train an SR model using the joint space of calibrated kernels and noise models and test it on data from a held-out device (\cref{subsec:experiments_cross_camera}).


\subsection{Implementation Details}
\label{subsec:implement_details}

We follow \cref{subsec:kernel_modeling}, to approximate 4$\times$ SR kernels. 
For each camera, we perform camera-display linearization and alignment and display $B=20$ random patterns.  To remove noise, we average a burst of 100 captures per displayed pattern to form $\mathbf{y}$. 
The FOV corresponding to 4$\times$ zoom is divided into patches of 128$\times$128, 
and for each RAW patch $\mathbf{y}$, we synthesize the corresponding 512$\times$512 HR RGB pattern, $\mathbf{x}$. 
We use a differentiable warping function in \cref{eq:monitor_to_sensor_formation_modeling} and the $\ell_1$-norm as $\mathcal{L(\cdot)}$ in \cref{eq:kernel_modeling}, which we optimize using ADAM~\cite{kingma2014adam}, 
to approximate $\mathbf{K}_r, \mathbf{K}_g,\mathbf{K}_b$ per patch.  

We calibrate noise parameters for a set of discrete ISO levels $\mathcal{S}=\{50, 100, 200, 400, 800, 1600, 3200\}$ per camera. 
We sample noise from a RAW burst of homogeneous regions in a color-checker chart to cancel out the potential impacts of non-uniform lighting. All camera sensors use Bayer CFAs, so we separately model noise for each of the four Bayer color channels. We fit a curve to each HG parameter along the calibrated ISOs, and interpolate when synthesizing random noise if the selected ISO $\notin \mathcal{S}$.

\subsection{Evaluation Metrics}
\label{subsec:experiments_metrics}


\paragraph{Reference-based evaluation} using metrics such as PSNR and SSIM is infeasible for real-world images due to the absence of ground-truth (GT) HR data. One possible workaround is to use a DSLR capture of the same scene as the reference, but this approach requires synchronized captures from two devices. Additionally, alignment poses a significant challenge, as HR images are often captured with a zoom lens that has a shallow depth of field, causing many regions to appear blurrier than their corresponding areas in the LR image.
Instead, we simplify this process by using a display prototype. We select high-quality DSLR captures from the MA5K dataset~\cite{fivek} and display them on a color-calibrated UHD monitor. The test camera then captures LR RAW images, and we align the displayed high-quality images with the RAW captures using optical flow and homography estimation while accounting for the SR scale (see supplemental material for details). Since the image is projected onto a display, creating a planar scene, alignment is more straightforward and less prone to errors.

\begin{figure*}[t]
\centering
\scriptsize
\noindent
\begin{minipage}{0.125\linewidth}
\centering
Original LR Capture 
\end{minipage}%
\begin{minipage}{0.125\linewidth}
\centering
4$\times$ Upsampling 
\end{minipage}%
\begin{minipage}{0.125\linewidth}
\centering
Bicubic Kernels
\end{minipage}%
\begin{minipage}{0.125\linewidth}
\centering
Kernel GAN~\cite{bell2019blind}
\end{minipage}%
\begin{minipage}{0.125\linewidth}
\centering
MANet~\cite{liang2021mutual}
\end{minipage}%
\begin{minipage}{0.125\linewidth}
\centering
Real-ESRGAN~\cite{wang2021realesrgan} 
\end{minipage}%
\begin{minipage}{0.125\linewidth}
\centering
Degradation-Transfer~\cite{chen2021extreme} 
\end{minipage}%
\begin{minipage}{0.125\linewidth}
\centering
Ours (Camera-specific)
\end{minipage}
\centering
\vspace{-0.1cm}
  \includegraphics[angle=0,valign=m,width=1\linewidth, trim={0 0 0cm 0.8cm},clip]{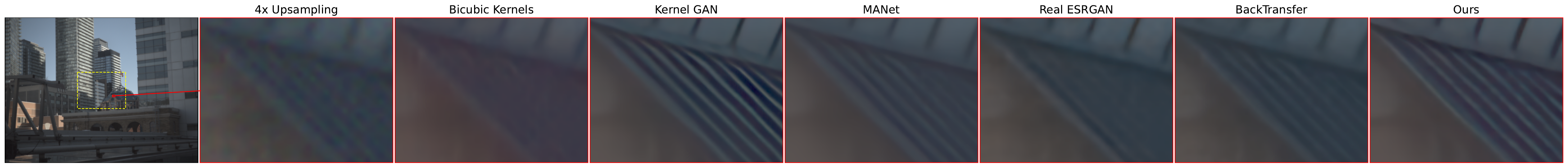}
  \vspace{-0.1cm}
 \includegraphics[angle=0,valign=m,width=1\linewidth, trim={0 0 0cm .8cm},clip]{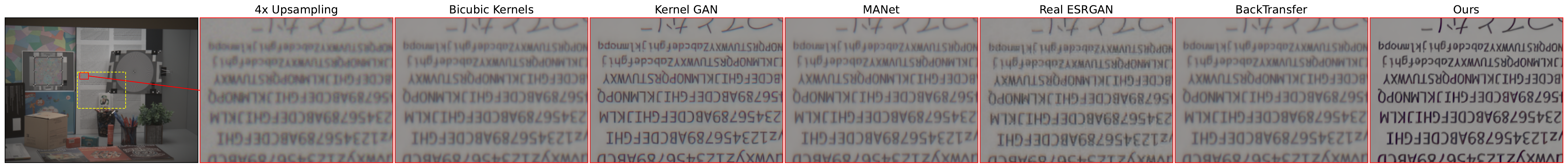} 
 \vspace{-0.1cm}
 \includegraphics[angle=0,valign=m,width=1\linewidth, trim={0 0 0cm 0.8cm},clip]{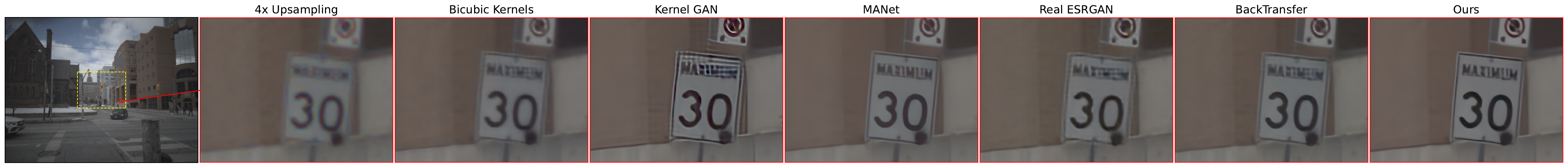}
 \vspace{-0.1cm}
 \includegraphics[angle=0,valign=m,width=1\linewidth, trim={0 0 0cm 0.8cm},clip]{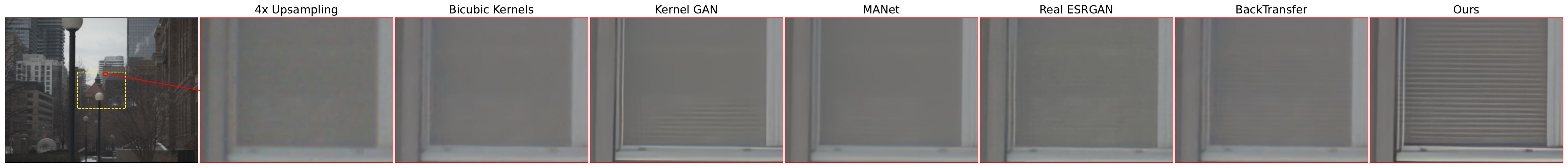} 
 \vspace{-0.3cm}
\caption{ \textbf{Camera-specific SR results} on data from four devices 
(from top to bottom: S23U Tele 1, S24U Tele 2, Pixel 9 Pro Main, and Pixel 9 Pro Tele).
We compare the outputs of a RAW-to-RGB 4$\times$ SR model trained on data synthesized with different degradation 
baselines and our own calibrated, camera-specific degradations. 
The model trained with our data produces sharper results, and recovers more structural details.
For better visualization, the original LR capture is linearly demosaicked and naively upsampled by 4$\times$. 
All images are white-balanced in each row using similar gains and gamma correction to ensure a consistent visual comparison.
The yellow dashed rectangle corresponds to the FOV of interest w.r.t 4$\times$ zoom factor. 
 }
\label{fig:qualitative_results_per_camera_models}
\end{figure*}

\begin{table*}[th]
    \centering
\footnotesize
\setlength\tabcolsep{1.5pt}     
\begin{tabular}{l|ccccc|ccccc|ccccc|ccccc}
\Xhline{2.0pt}
\multicolumn{1}{c|}{}                                  & \multicolumn{5}{c|}{S23U Tele 1}                                           & \multicolumn{5}{c|}{S24U Tele 2}                                           & \multicolumn{5}{c|}{Pixel 9 Pro Main}                                      & \multicolumn{5}{c}{Pixel 9 Pro Tele}                                      \\ \cline{2-21} 
                                                Method       & \multicolumn{2}{c}{No Ref. Metrics} &  & \multicolumn{2}{c|}{Ref. Metrics} & \multicolumn{2}{c}{No Ref. Metrics} &  & \multicolumn{2}{c|}{Ref. Metrics} & \multicolumn{2}{c}{No Ref. Metrics} &  & \multicolumn{2}{c|}{Ref. Metrics} & \multicolumn{2}{c}{No Ref. Metrics} &  & \multicolumn{2}{c}{Ref. Metrics} \\ \cline{2-3} \cline{5-8} \cline{10-13} \cline{15-18} \cline{20-21} 
                                                 & MTF50            & MTF25            &  & PSNR            & SSIM            & MTF50            & MTF25            &  & PSNR            & SSIM            & MTF50            & MTF25            &  & PSNR            & SSIM            & MTF50            & MTF25            &  & PSNR            & SSIM           \\ \cline{1-18} \cline{20-21} 
Bicubic                                                & 0.05             & 0.29             &  & 32.01           & 0.952           & 0.16             & 0.28             &  & 33.42           & 0.935           & 0.54             & 0.95             &  & 30.88           & 0.874           & 0.08             & 0.22             &  & 30.96           & 0.868          \\
KernelGAN~\cite{bell2019blind}        & 1.42             & 1.15             &  & 33.12           & 0.950           & 1.62             & 1.35             &  & 33.48           & 0.937           & 1.64             & 1.05             &  & 31.10           & 0.856           & \textbf{1.46}    & 1.22             &  & 32.66           & 0.883          \\
MANet~\cite{liang2021mutual}          & 0.31             & 0.89             &  & 33.42           & 0.959           & 0.29             & 0.48             &  & 33.47           & 0.935           & 1.66             & 1.24             &  & 33.14           & 0.889           & 0.38             & 0.62             &  & 32.93           & 0.886          \\
Degradation-Transfer~\cite{chen2021extreme}   & 1.31             & 0.94             &  & 33.47           & 0.954           & 1.56             & 1.29             &  & 33.01           & 0.948           & 1.61             & 1.12             &  & 32.98           & 0.885           & 1.18             & 0.68             &  & 32.76           & 0.878          \\
Real-ESRGAN~\cite{wang2021realesrgan} & 0.72             & 0.83             &  & 32.92           & 0.941           & 1.30             & 1.00             &  & 33.03           & 0.949           & 1.14             & 0.83             &  & 33.38           & 0.888           & 1.15             & 0.62             &  & 33.21           & \textbf{0.889} \\
Ours (Camera-specific)                                                   & \textbf{1.50}    & \textbf{1.18}    &  & \textbf{33.59}  & \textbf{0.961}  & \textbf{2.00}    & \textbf{1.63}    &  & \textbf{33.53}  & \textbf{0.956}  & \textbf{1.72}    & \textbf{1.20}    &  & \textbf{33.47}  & \textbf{0.901}  & 1.45             & \textbf{1.23}    &  & \textbf{33.32}  & \textbf{0.889} 
\\
\Xhline{2.0pt}
\end{tabular}\vspace{-0.3cm}
    \caption{
    Quantitative \textbf{camera-specific SR results} on data captures with four different cameras.
    The 4$\times$ SR model trained with our calibrated, camera-specific degradations outperforms models trained with the baseline degradation pools, in terms of all metrics used.}
    \label{tab:per_cam_degrdation_mdel} \vspace{-0.3cm}
\end{table*}

\begin{table}[t]
    \centering
\footnotesize
\setlength{\aboverulesep}{0mm}
\setlength{\belowrulesep}{0mm}
\renewcommand{\arraystretch}{1.0}
\setlength\tabcolsep{0pt}  
\begin{tabular}{
p{0.21\linewidth} 
p{0.31\linewidth} 
>{\centering}p{0.12\linewidth} 
>{\centering}p{0.12\linewidth} 
>{\centering}p{0.12\linewidth} 
>{\centering\arraybackslash}p{0.12\linewidth} 
}
\Xhline{2.0pt}
\multirow{2}{*}[-0.5\dimexpr \aboverulesep + \belowrulesep + \cmidrulewidth]{\makecell{Camera}}
&\multirow{2}{*}[-0.5\dimexpr \aboverulesep + \belowrulesep + \cmidrulewidth]{\makecell{Method}}
& \multicolumn{2}{c}{No Ref. Metrics} & \multicolumn{2}{c}{Ref. Metrics} \\
\cmidrule{3-4} \cmidrule{5-6} 
               &                                                                                    & MTF50           & MTF25           & PSNR            & SSIM           \\ \hline

\multirow{4}{*}[-0.5\dimexpr \aboverulesep + \belowrulesep + \cmidrulewidth]{\makecell{Pixel 6 Main}} 
& Real-ESRGAN~\cite{wang2021realesrgan}        &               0.38	 &   0.50                 & 31.75               & 0.884      \\
& BSRAW~\cite{conde2024bsraw} &                               0.97 &	0.82 & 28.79               & 0.713                \\
& RAWSR~\cite{xu2019towards}                           &          0.96	& \textbf{0.90}      & 28.99              & 0.790            \\         
                     & Ours (Cross-camera)                                     &     \textbf{1.19}             &     0.86             & \textbf{32.41}               & \textbf{0.905}      \\ 
\hline
\multirow{4}{*}[-0.5\dimexpr \aboverulesep + \belowrulesep + \cmidrulewidth]{\makecell{Mi11 Main}}
 & Real-ESRGAN~\cite{wang2021realesrgan}                                        &      0.33            &         0.40         &   35.08             &  0.914           \\
& BSRAW~\cite{conde2024bsraw}  &        0.34 & 0.43       &  33.65              &  0.817  \\
& RAWSR~\cite{xu2019towards}                            &      0.14	& 0.22      & 35.71   & 0.914                       \\
                     & Ours (Cross-camera)                                &                \textbf{0.42}	& \textbf{0.59}                & \textbf{36.11}               & \textbf{0.919}          \\                     
\Xhline{2.0pt}                     
\end{tabular}
\vspace{-3mm}
    \caption{Quantitative \textbf{cross-camera SR results} on data captures with Pixel 6  Main and Mi 11 Main RAW cameras.
    %
    %
    Our 4$\times$ SR model is trained on synthetically generated data using degradation models obtained from seven different cameras. The two test camera degradations are not seen by any of the models during training.}
    \label{tab:blind_SR_baselines}
    \vspace{-3mm}
\end{table}

\paragraph {No-reference evaluation.} 
We use the modulation transfer function (MTF) to evaluate how well the SR model recovers fine details. MTF is computed by comparing image contrast to a reference target contrast. To measure contrast and frequency loss across the FOV, we use a 3$\times$6 grid of a Siemens star pattern with 20 spokes~\cite{loebich2007digital}. The spatial frequencies at which relative contrast values reach 25\% and 50\% are labeled as MTF25 and MTF50, respectively. These metrics serve as indicators of the SR model’s ability to restore scene details. Additional details are in the supplemental material.

\subsection{Baselines}
\label{subsec:baselines}

\noindent\textbf{Bicubic Kernels}:
We use 4$\times$ bicubic down-sampling kernels as the most common approach to obtain LR images.

\noindent\textbf{KernelGAN}~\cite{bell2019blind}:
We use the official implementation
to extract 4$\times$ blur kernels from smartphone images. 
Because these images are high-resolution (typically $~$12MP), we split them into non-overlapping 480$\times$480 patches 
and compute a kernel for each patch.
We visually inspect the computed kernels and discard those that do not resemble a valid PSF, ultimately retaining approximately 400 kernels per device.

\noindent\textbf{MANet}~\cite{liang2021mutual} enables dense, pixel-wise kernel estimation. However, applying it directly to the HR images in our experiments would generate an impractically large number of kernels with significant redundancy. To manage this, we extract 480$\times$480 patches and use MANet to estimate kernels, then randomly sample a subset of 1000 kernels per patch. This results in approximately 400,000 kernels per device, from which we randomly select 400 for use.

\noindent\textbf{Real-ESRGAN}~\cite{wang2021realesrgan}:
We follow the official settings, applying random isotropic and anisotropic Gaussian blur, random resizing (nearest, bilinear, and bicubic), and random Gaussian and Poisson noise to linearized RGB HR images. JPEG compression is omitted, as it is unsuitable for linear inputs. Mosaicking is performed last to generate the RAW input.

\noindent\textbf{Degradation-Transfer}~\cite{chen2021extreme}
uses captures of designated patterns to obtain paired data for estimating lens PSFs (\ie 1$\times$ SR kernels). 
We adapt this approach for 4$\times$ SR kernel estimation on the same patches of the FOV used in \cref{subsec:implement_details}. 

\noindent\textbf{RAWSR}~\cite{xu2019towards}:
We generate disk and motion blur kernels using the same radius and kernel size ranges as described in the original paper. Since the application of the AHD demosaicking algorithm to degraded RAW LR images ($X_{ref}$ in \cite{xu2019towards}) during training is not explicitly detailed, we instead produce $X_{ref}$ for this baseline using DCRAW demosaicker.

\noindent\textbf{BSRAW}~\cite{conde2024bsraw}:
We follow their implementation details and use a NAFNet-based network~\cite{chen2022simple} for RAW-to-RAW SR. The training incorporates their diverse pool of blur kernels and sensor noise models, including those from~\cite{SIDD_2018_CVPR}.
Since this baseline focuses on RAW-to-RAW SR, we apply a CFA-like sub-sampling to the HR images to generate training targets. During inference, the outputs are processed with DCRAW to obtain RGB results~\cite{conde2024bsraw}.

\subsection{Camera-specific SR Model Assessment}
\label{subsec:experiments_per_camera}
We synthesize RAW data from 256$\times$256 patches, extracted from the 900 training images in the DIV2K dataset~\cite{Agustsson_2017_CVPR_Workshops}. We train a SR model with a degradation pool that includes only our camera-specific RGB kernels and noise, as well as a SR model per degradation baseline with the same HR data. 

For the SR network, we use the RDBNet architecture~\cite{wang2018esrgan}. Some baselines (\ie, KernelGAN and MANet) are designed to approximate kernels in the RGB domain, so we precede our model with a differentiable, fixed linear demosaicker to prevent a domain gap. Additionally, we use the $\ell_1$-norm as the loss function, omitting GAN and perceptual loss terms to avoid hallucination artifacts.

For \emph{qualitative} evaluation, we capture a set of 30 RAW images with each camera, including outdoor and indoor scenes as well as 30 RAW-GT paired data, obtained using the display prototype (\cref{subsec:experiments_metrics}) for \emph{quantitative} analysis. 
Additionally, we capture a RAW image of the Siemens star pattern for MTF25 and MTF50 calculations. \cref{fig:qualitative_results_per_camera_models} presents qualitative results of RAW-to-RGB SR on S23U Tele 1, S24U Tele 2, Pixel 9 Pro Main, and Pixel 9 Pro Tele RAW captures. PSNR, SSIM, and MTF scores for the four cameras are listed in \cref{tab:per_cam_degrdation_mdel}. Our proposed degradation framework for camera-specific modeling consistently improves performance across all metrics.
%
Note that, \eg, a relative MTF50 of 2.00 in the S24U Tele 2 experiment indicates that our camera-specific 4$\times$ SR model increases the spatial frequency at which the imaging system can reproduce $50\%$ of the contrast in detail by $200\%$; using MANet kernels, the improvement is only $29\%$. This enhanced detail recovery is reflected in the reduction of ringing, color-bleeding, and over/under-shoot artifacts, shown in \cref{fig:qualitative_results_per_camera_models}.

\subsection{Cross-camera SR Model Assessments}
\label{subsec:experiments_cross_camera}

%
We extract 256$\times$256 patches from the training subset of the MA5K dataset~\cite{fivek} 
to synthesize LR-HR pairs for training. 
MA5K consists of six sets, each containing 5,000 images. 
The first set includes high-quality RAW images in DNG format, captured by a variety of DSLR cameras, while the remaining five sets contain RGB images derived from the original DNGs, each color-retouched by five different experts. 
In our case, we only use the RAW images, rendered into linear RGB. 
Although RAW images capture high-fidelity details, noise can be introduced during the capturing process due to varying ISO ranges, negatively impacting the quality of the rendered HR RGB images. To address this, we first apply RAW denoising using Adobe Lightroom, followed by demosaicking with DCRAW. The resulting linear RGB images are used as ground-truth, and LR RAW images are then generated, as described in \cref{subssec:method_synthesizing_traing_data}.

We employ the same RRDBNet architecture used in \cite{wang2018esrgan} as our SR network, with some modifications 
for RAW-to-RGB SR.
Input RAW data are converted into 4-channel GBRG images by stacking pixels in every 2$\times$2 block into separate color channels. 
To account for the 2$\times$ reduction in image resolution caused by such a conversion, we add an up-sampling layer before the final convolutional block. 
We use the $\ell_1$-norm as the loss function, as in \cref{subsec:experiments_per_camera}.

\begin{figure*}[t]
\centering
\footnotesize
\noindent
\begin{minipage}{0.166\linewidth}
\centering
Original LR Capture 
\end{minipage}%
\begin{minipage}{0.166\linewidth}
\centering
4$\times$ Upsampling 
\end{minipage}%
\begin{minipage}{0.166\linewidth}
\centering
Real-ESRGAN~\cite{wang2021realesrgan} 
\end{minipage}%
\begin{minipage}{0.166\linewidth}
\centering
BSRAW~\cite{conde2024bsraw}
\end{minipage}%
\begin{minipage}{0.166\linewidth}
\centering
RAWSR~\cite{xu2019towards} 
\end{minipage}%
\begin{minipage}{0.166\linewidth}
\centering
Ours (Cross-camera)
\end{minipage}
\centering
\vspace{-0.1cm}
 \includegraphics[angle=0,valign=m,width=1\linewidth, trim={0 0 0cm 0.8cm},clip]{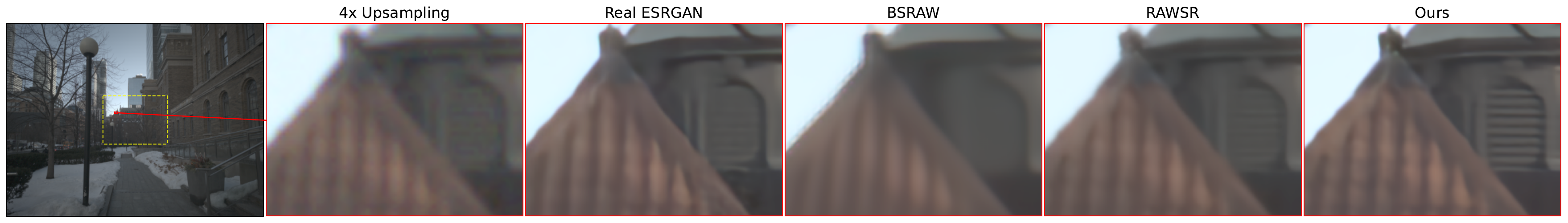}
 \vspace{-0.1cm}
  \includegraphics[angle=0,valign=m,width=1\linewidth, trim={0 0 0cm 0.8cm},clip]{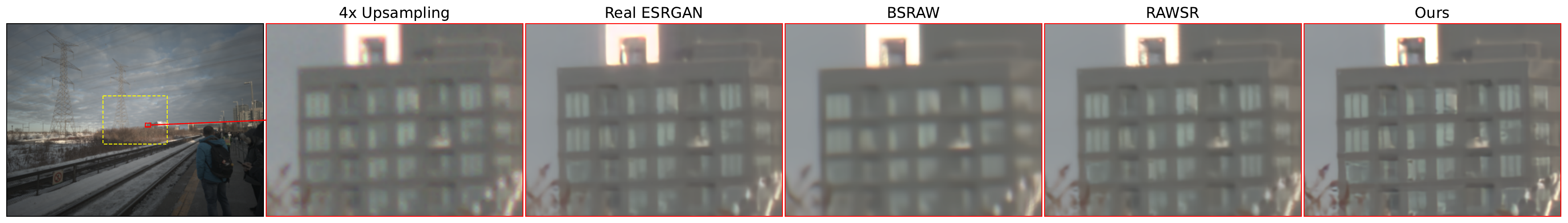} 
 \vspace{-0.1cm}
 \includegraphics[angle=0,valign=m,width=1\linewidth, trim={0 0 0cm 0.8cm},clip]{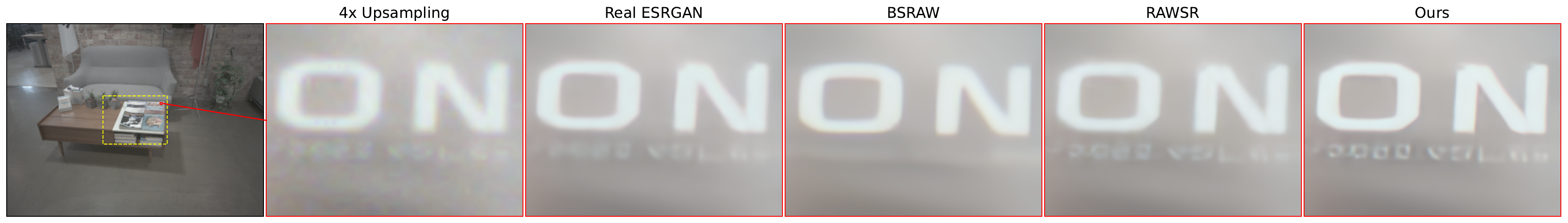} 
 \vspace{-0.3cm}
 \caption{\textbf{Cross-camera SR results on Mi 11 Main real RAW captures}. 
 We compare the outputs of a RAW-to-RGB $4\times$ SR model, trained on data generated with different degradation modeling approaches.
 The degradations of the test camera are not seen by any of the models during training, yet
 the model trained with our synthetic data is better at recovering fine details such as building structures and text.
 The original LR capture is linearly demosaicked, and naive $4\times$ upsampling is applied for visualization.  
 All images are white-balanced with similar gains, and gamma corrected. The yellow dashed rectangle represents the FOV of interest relative to the $4\times$ zoom factor.}
 \vspace{-0.3cm}
\label{fig:qualitative_results_Xiaomi_models}
\end{figure*}

\begin{figure}[t]
    \centering
\includegraphics[width=1.0\linewidth,trim={0 0.5cm 0cm 0.0cm},clip]{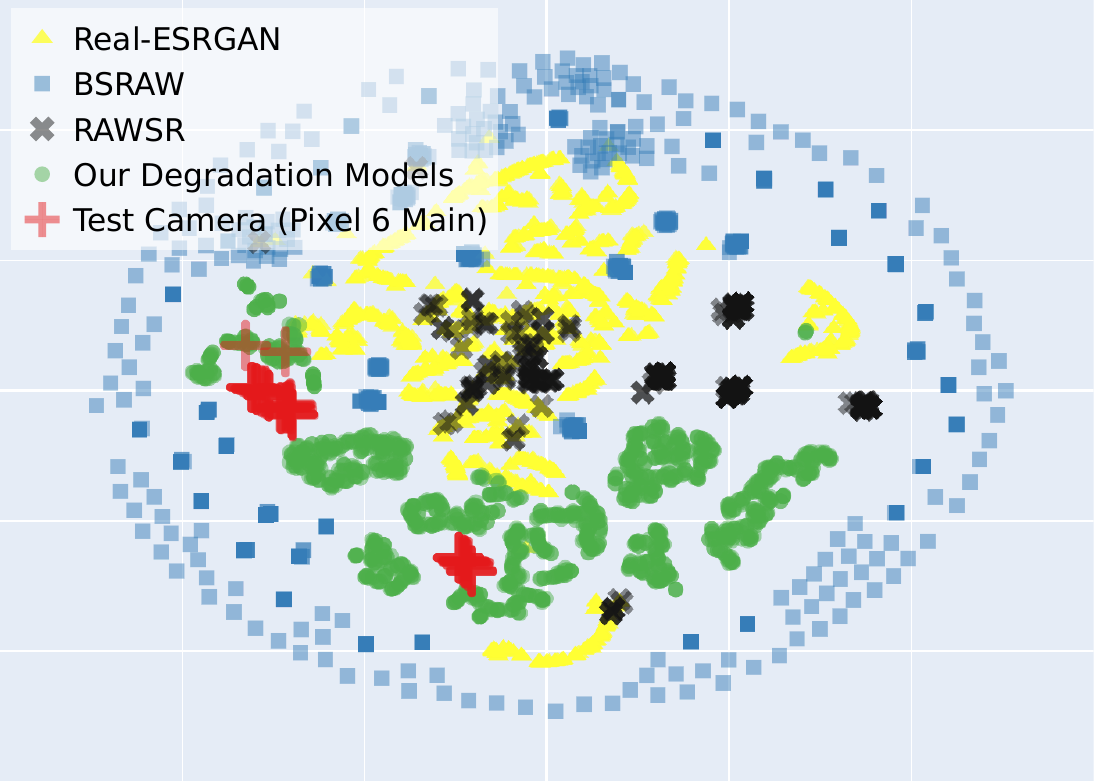}%
\vspace{-0.2cm}\caption{Distributions of calibrated 4$\times$ SR kernels for Pixel 6 Main and kernels from other degradation protocols. 
Pixel 6 kernels lie closer to those calibrated for other devices using our framework, hinting at similarities 
between RAW domains of different smartphones. 
For this visualization, we first run PCA~\cite{pearson1901liii} to reduce the dimension of 21$\times$21 kernels to 50 followed by t-SNE~\cite{van2008visualizing} to further reduce dimensions to 2. 
%
}\vspace{-0.3cm}
\label{fig:kernels_pixel6_distribution}
\end{figure}

We evaluate our model on RAW images captured by Mi 11 Main and Pixel 6 Main, in a similar fashion as in 
\cref{subsec:experiments_per_camera}. 
\cref{fig:qualitative_results_Xiaomi_models,fig:teaser} compare qualitative results of RAW to RGB for 4$\times$ SR models trained using our degradations to other baselines. 
Note that the SR kernels and noise models specific to these cameras are excluded from the degradation pool during training data synthesis to mimic a blind SR case.
\cref{tab:blind_SR_baselines} lists PSNR, SSIM, as well as MTF50 and MTF25 scores. 
The model trained with our degradations outperforms all other baselines on unseen data by a significant margin, demonstrating the practical benefits of accurate degradation modeling in a real-world use case. 
 
To further illustrate that degradations modeled using our framework are more ``similar'' to those corresponding to the held-out devices, we plot the distribution of SR kernels from various baselines, using t-SNE embeddings~\cite{van2008visualizing}, in \cref{fig:kernels_pixel6_distribution}. 
In the same plot, we visualize SR kernel embeddings modeled using our framework, for the seven smartphone cameras used during training, and the held-out Pixel 6 Main.

\section{Conclusions and Limitations}
\label{sec:conclusions}
We have described a calibration-based framework for modeling degradation blur and sensor noise in smartphone cameras, specifically for super-resolution (SR). We have used our proposed framework to generate paired synthetic LR-HR data from sRGB images, in an ``unprocessing'' fashion, and used them to train RAW-to-sRGB SR models that are evaluated on real smartphone captures. Our experiments suggest that accurate degradation modeling produces synthetic data that are more faithful to real RAW, which in turn improves the performance of SR models in real-world use cases. More importantly, our approach reduces the domain gap between synthetic and real RAW, 
making the SR model capable of better generalizing to similar smartphone sensors. We have demonstrated the advantages of our approach through qualitative and quantitative comparisons,both in the camera-specific (non-blind SR) and the camera-agnostic (blind SR) setup. SR models trained with our improved synthetic data significantly outperform baselines that employ simple parametric degradation models, in terms of PSNR, SSIM, and MTF scores, while also exhibiting fewer visual artifacts. 
While our kernel and noise calibration requires a specialized capture setup and is somewhat laborious, the resulting calibration data generalizes well to other mobile cameras.
Moreover, our conclusions are only applicable to smartphones and under good lighting conditions. Although we do not expect them to generalize to more complicated DSLR cameras, or in low-light conditions, this would be an interesting direction for future work. In the meantime, we will publicly release our calibrated blur kernels and noise models, to facilitate further research on this topic.

 \clearpage

{
    \small
    \bibliographystyle{ieeenat_fullname}
    \bibliography{digital_zoom}
}

\clearpage


\renewcommand{\thefigure}{S\arabic{figure}}
\renewcommand{\thesection}{S\arabic{section}}
\renewcommand{\thetable}{S\arabic{table}}
\renewcommand{\thelstlisting}{S\arabic{lstlisting}}
\renewcommand{\theequation}{S\arabic{equation}}

\setcounter{figure}{0}
\setcounter{table}{0}
\setcounter{section}{0}
\setcounter{equation}{0}

\maketitlesupplementary

This supplementary material contains additional details regarding the proposed degradation modelling for RAW SR method and the experimental assessments we presented in the main document.

\section{RAW Capturing Setup}
\label{sec:capture_setup}

\begin{figure}[t]
\centering
    \includegraphics[angle=0,valign=m,width=0.95\linewidth, trim={0 0 0cm 0},clip]{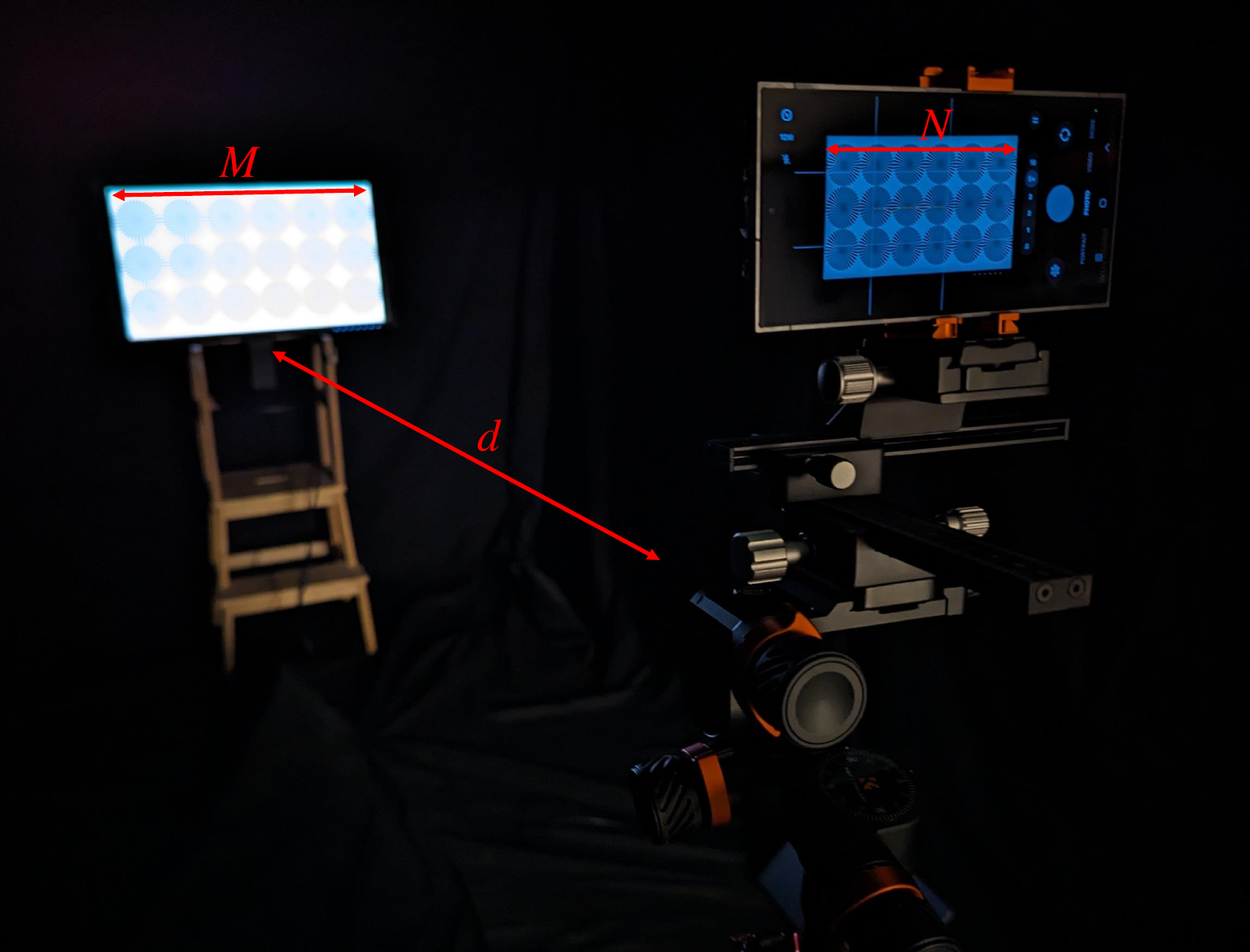} 
    \vspace{-0.3cm}
    \caption{Our calibration setup and display prototype.}
    \vspace{-6pt}
\label{fig_sup:calibration_setup}
\end{figure}

\cref{fig_sup:calibration_setup} shows a snapshot of our capturing setup, which consists of a smartphone mounted on a tripod that captures an image displayed on the display prototype. This system is used to collect images for kernel modeling (\cref{subsec:kernel_modeling} of the main paper) and paired LR-HR images for reference-based evaluation metrics (\cref{subsec:experiments_metrics}).
%

\vspace{-6pt}
\paragraph{Camera-display framing.}
To effectively model subsampling performed in the camera imaging pipeline during the kernel modeling and evaluation data collection, 
the target display and camera are framed such that the spatial resolution of the displayed image exceeds the target resolution of the sensor. 
To this end, we display a HR image on the monitor with 1:1 scaling to prevent interpolation or stretching in the display domain. The distance $d$ between the display and camera is adjusted based on the target SR scale.

Specifically, the scaling factor $s$ is defined as the ratio between the displayed image resolution $M$ and the captured resolution $N$, where $M$ corresponds to the spatial resolution of the HR image displayed on the monitor, and $N$ corresponds to the resolution captured by the camera sensor.  The distance $d$ is then determined such that $s=\frac{f}{d}$, where $f$ is the focal length of the camera. By adjusting the distance $d$, the camera captures the displayed image at the target resolution, simulating the effect of subsampling and ensuring the captured data is consistent with the intended target scale for kernel modeling and evaluation. This approach helps avoid unwanted interpolation and aliasing, providing more accurate data that reflects real-world imaging scenarios.

\vspace{-6pt}
\paragraph{Camera specifications.}
We provide a list of the mobile cameras used in our experiments (\cref{sec:experiments}). As shown in the table, these cameras have varying optical properties and sensor sizes, resulting in diverse degradation kernels and noise profiles.
Cameras that are common across multiple devices are excluded. For example, the Main and Tele 1 cameras of Samsung S24U  are identical to those in S23U Main and Tele 1, making separate degradation modeling unnecessary. 
Additionally, ultra-wide cameras are excluded from the degradation modeling, as they are not commonly used for digital zoom applications.

\vspace{-6pt}
\paragraph{RAW capture application.}
The native camera applications on mobile phones do not necessarily provide direct access to RAW captures. For obtaining RAW data, a dedicated RAW capturing tool is required that meets the following criteria: (1) RAW access of the cameras listed in \cref{sec:experiments} and \cref{tab:list_of_cameras}, (2) Manual ISO, shutter speed, and focus controls, and (3) Burst imaging capability in RAW format. While burst capturing can be simulated by sequentially triggering the shutter to capture a scene, this may cause misalignment between the captures due to random spatial shifts caused by the optical image stabilizer (OIS), even in vibration-free remote capturing setups.
For RAW data capturing, including noise calibration, kernel modeling, and evaluation data collection, we use the Android ProShot, which meets the above requirements. The capture process is automated through android debug bridge (ADB) commands, enabling remote control of cameras.

\begin{table}[t]
    \centering
\footnotesize
\setlength\tabcolsep{1.0pt}      
\begin{tabular}{lccccc}
\Xhline{2.0pt}
Camera           &  \begin{tabular}[c]{@{}c@{}}Optical\\ Zoom\end{tabular}& \begin{tabular}[c]{@{}c@{}}Focal\\ Length\end{tabular} & \begin{tabular}[c]{@{}c@{}}Sensor\\ Size\end{tabular} & \begin{tabular}[c]{@{}c@{}}Pixel\\ Size\end{tabular}  & \begin{tabular}[c]{@{}c@{}}Sensor\\ Resolution\end{tabular} \\
\hline
S23+ Main        & 1x           & 23 mm        & 1/1.56''     & 1.0 $\mu$m     & 4080x3060 (50MP)  \\
S23+ Tele        & 3x           & 69 mm        & 1/3.94''     & 1.0 $\mu$m     & 3648x2736         \\
S23U Main        & 1x           & 24 mm        & 1/1.3''      & 0.6 $\mu$m     & 4080x3060 (200MP) \\
S23U Tele 1      & 3x           & 69 mm        & 1/3.52''     & 1.12 $\mu$m    & 3648x2736         \\
S24U Tele 2      & 5x           & 111 mm       & 1/2.52''     & 0.7 $\mu$m     & 4080x3060 (50MP)  \\
Pixel 6 Main     & 1x           & 26 mm        & 1/1.31''     & 1.2 $\mu$m    & 4080x3072 (50MP)  \\
Pixel 9 Pro Main & 1x           & 25 mm        & 1/1.31''     & 1.2 $\mu$m     & 4080x3072 (50MP)  \\
Pixel 9 Pro Tele & 5x           & 113 mm       & 1/2.55''     & 0.7$\mu$m      & 4032x3024 (48MP) 
\\
Mi 11 & 1x           & 25 mm       & 1/1.52''     & 0.7$\mu$m      & 6016x4512 (108MP) 
\\
\Xhline{2.0pt}
\end{tabular}
    \vspace{-3pt}
    \caption{Specifications~\cite{PhoneSpecs} of all the cameras used in our experiments. 
    Note that some of the sensors are designed for binning and Android ProShot cannot access the unbinned version of the RAW captures. Sensor Resolution column indicates the captured RAW resolution in binned (regular Bayer) format, while the number in parentheses indicates the actual sensor resolution before binning.}
    \label{tab:list_of_cameras}
    \vspace{-8pt}
\end{table}

\begin{figure}[t]
\centering
\includegraphics[angle=0,valign=m,width=0.8\columnwidth, trim={0 0 0cm 0},clip]{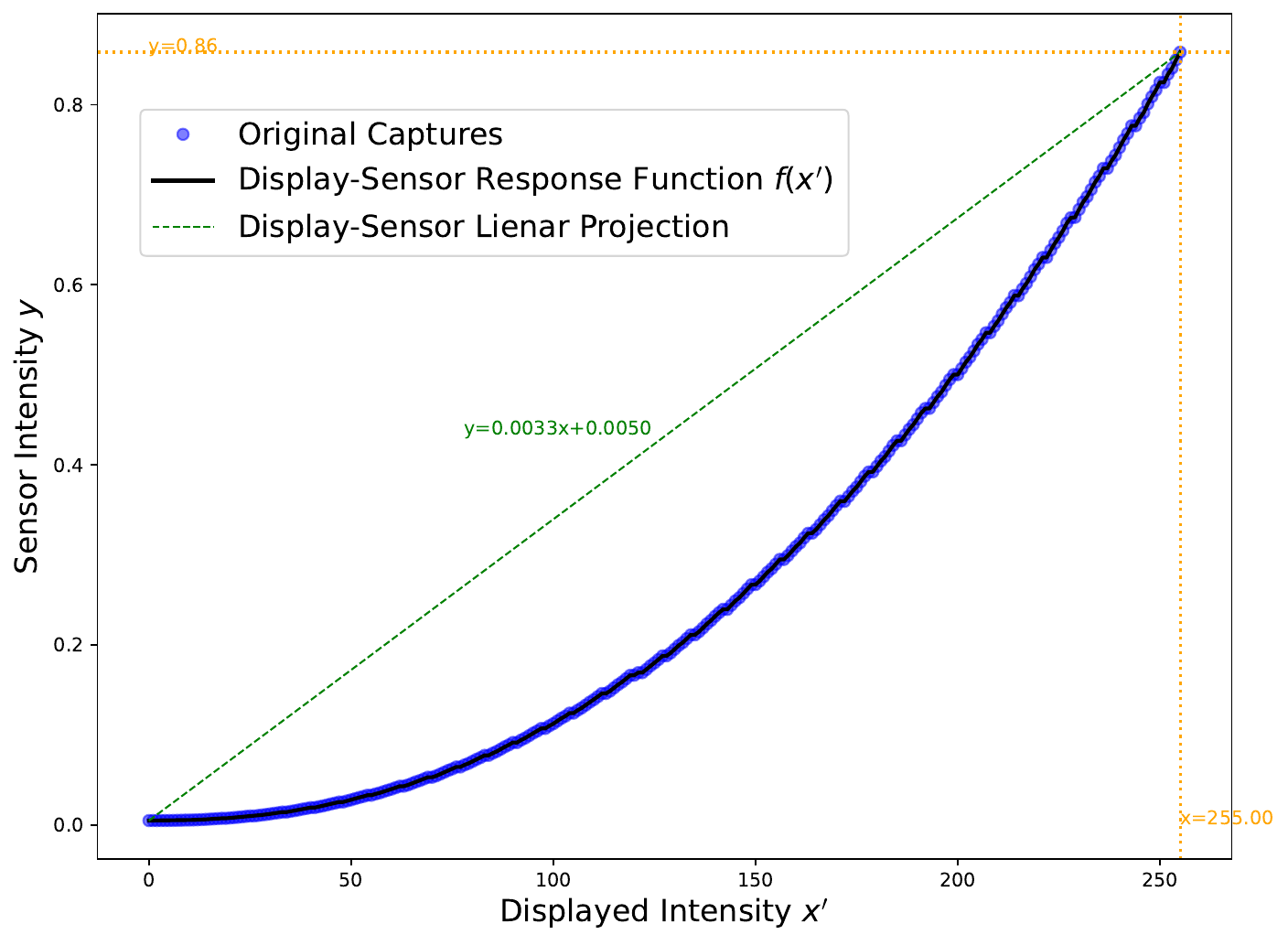} 
 \vspace{-0.3cm}
\caption{Display-to-sensor linearization function obtained for our display prototype and Pixel 9 Pro Main camera.}
\label{fig_sup:monitor_curve}
\end{figure}

\section{Radiometric Alignment between HR-LR}
\label{sec:GT_color_and_linearization}
We employ a display prototype to generate HR-LR image pairs in both SR kernel modeling (Sec.~\ref{subsec:kernel_modeling}) and SR model evaluation using reference-based metrics (Sec.~\ref{subsec:experiments_metrics}).
%
The displayed image undergoes a non-linear transformation on the display, such as tone mapping or gamma correction, before being captured on the target camera sensor. 

To determine the non-linear mapping from the displayed image to sensor space, we first display gray patches at 255 different steps (gray-scale 8-bit format images, from 0 to 255), \ie $\text{max}_{ \acute{\mathbf{x}}} = \text{255}$ and capture a RAW image for each displayed patches. The average intensity of the captured gray patches and their corresponding values displayed values are used to fit a non-linear curve to the measurements, denoted as $f(\acute{\mathbf{x}})$. The display-to-sensor response curve obtained using this process for the Pixel 9 Pro Main camera is shown in \cref{fig_sup:monitor_curve}.
We then display 140 color patches corresponding to the color values found in ColorChecker Digital SG. Using the values obtained for the color patches in the senor domain and their linearized displayed sRGB values \ie $f^{-1}(\acute{\mathbf{x}})$, we compute a 3$\times$3 color correction matrix (CCM). This CCM, denoted by $\mathbf{C}$, together with the inverse of the monitor-to-sensor response curve, forms our color mapping function from displayed domain to the sensor color space, given by:
\begin{align}
\Linearize(\acute{\mathbf{x}}) = \mathbf{C} f^{-1}(\acute{\mathbf{x}}).
\end{align}

\section{Kernel Modeling More Details}
\label{sec:kernel_setup_alignemnt}

\begin{figure}[t]
\centering
 \includegraphics[angle=0,valign=m,width=0.99\linewidth, trim={3.7cm 3.7cm 3.7cm 3.7cm},clip]{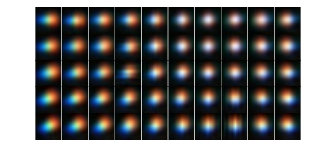} 
 \vspace{-0.3cm}
\caption{Examples of camera-specific SR kernels we calibrate and add to the pool of SR degredations; Calibrated 4$\times$ SR kernels for the center of FOV of S23P Tele camera.
}
\label{fig_sup:s23p_3x_kernels_vilsualization}
\end{figure}

In \cref{fig_sup:s23p_3x_kernels_vilsualization}, we show examples of kernels for 4$\times$ SR that we calibrate for S23P Tele camera. Below, we describe additional details regarding the kernel calibration process.

\vspace{-6pt}
\paragraph{Sensor-to-display alignment.}
The camera and target display are geometrically aligned to approximate a perspective homography, which is used to initialize $\mathbf{H}$ in \cref{eq:kernel_modeling}. For geometric alignment, we use gray-code structured patterns (\cref{fig:monitor_sensor_image_fromation_examples} in the main paper) to establish the mapping between pixels in the display and their corresponding pixels in the senor~\cite{sels2019camera}. 
We employ a sequence of 23 vertical grey-code stripes and 23 horizontal grey-code stripes to encode the pixel locations on the screen and identify the corresponding positions on the sensor. These dense pairs of corresponding locations are then used to fit a homography matrix $\mathbf{H}$, which facilitates mapping of any displayed pattern on the monitor to the sensor domain as $\Warp(\acute{\mathbf{x}}, \mathbf{H})$ in \cref{eq:monitor_to_sensor_formation_modeling}.  

We follow the steps outlined in \cref{sec:GT_color_and_linearization} to calibrate the non-linear color space of the display and the color transformation from the displayed domain to the sensor. However, when the displayed patterns consist only black-and-white pixel values, the linearization and color mapping can be performed more efficiently.

Let $\mathbf{y}_1$ and $\mathbf{y}_2$ denote RAW captures of a pair of white ($\max_{ \acute{\mathbf{x}}}$) and black ($0$) patches $\mathbf{x}_1$ and $\mathbf{x}_2$ displayed on the monitor, respectively (\cref{fig:monitor_sensor_image_fromation_examples})). 
Then, the linearization and color mapping can be simplified to:
\begin{align}
& \Linearize \left( \mathbf{x} \right) = \frac{ \left(  \frac{ \mathbf{x} \left( \text{max}_{ \mathbf{y}} - \text{min}_{ \mathbf{y}} \right)}{ \text{max}_{ \acute{\mathbf{x}}} }  +  \text{min}_\mathbf{y} \right)  - b } {w - b},
\end{align}
where $\max_{ \mathbf{y}}=\text{Mean} \left(\mathbf{y}_1 \right)$ and $ \min_{ \mathbf{y}}=\text{Mean} \left(\mathbf{y}_2 \right)$ are mean intensities of RAW captures of a pair of white and black patches $\mathbf{x}_1$ and $\mathbf{x}_2$ displayed on the monitor, and $w$ and $b$ denote the white level and the black level of the sensor, respectively. Here, $\text{Mean}(\cdot)$ denotes spatial averaging.

Using the white and black field captures $\mathbf{y}_1$ and $\mathbf{y}_2$, we can model the lens vignette effect combined with the non-uniformity of the brightness on the monitor. It is straightforward to show that the combination of $\mathbf{v}$ and $\Linearize(\cdot)$ can be modeled as: 
\begin{align}
& \mathbf{v} \odot  \Linearize  \left( \mathbf{x} \right) = \frac{ \left(  \frac{ \mathbf{x} \odot  \left( \mathbf{y}_1 -\mathbf{y}_2 \right)}{ \text{max}_{ \acute{\mathbf{x}}} }  +  \mathbf{y}_2 \right)  - b } {w - b},
\end{align}
where $\odot$ denotes element-wise multiplication. As seen here, using only two pixel intensities--$0$ and $\max_{ \acute{\mathbf{x}}}$--to design kernel modeling patterns makes the model less sensitive to the non-linear color space of the display. 


\vspace{-6pt}
\paragraph{Capture details.}
Our monitor-to-sensor image formation model in \cref{eq:monitor_to_sensor_formation_modeling} assumes that the observations are noise-free. Thus, we capture a burst of 100 images of the displayed pattern at the lowest ISO level available in the ProShot RAW-capturing application and average the images to obtain $\mathbf{y}$ as a noise-free RAW image for \cref{eq:kernel_modeling}.

The exposure time is adjusted so that the white pixels displayed on the monitor map to approximately $80\%$ of sensor' white-level. This ensures good contrast in the captures while avoiding saturation of the pixels.




\begin{figure}[t]
\centering
\includegraphics[angle=0,valign=m,width=0.6\linewidth, trim={0 0 0cm 0},clip]{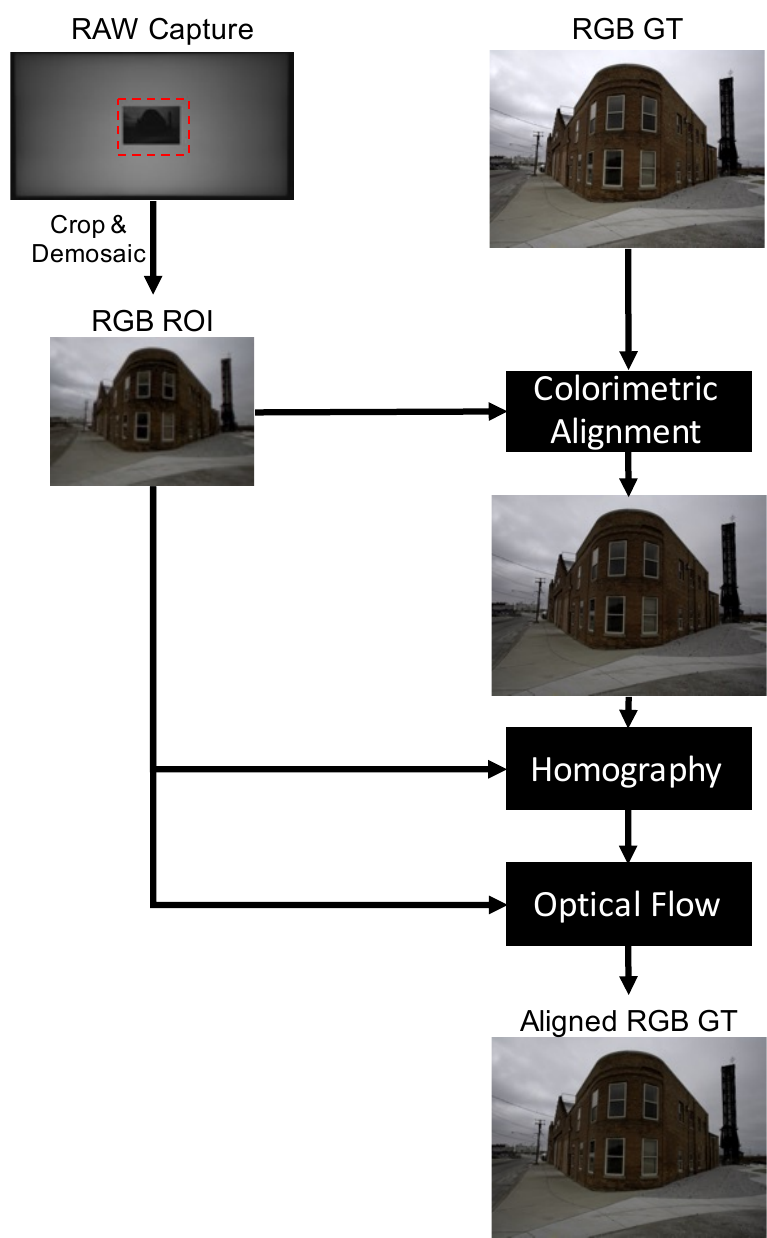}
\vspace{-0.2cm}
\caption{Our alignment method for producing paired evaluation data. After aligning the colour and brightness of the ground-truth image with a demosaiced crop of the monitor capture, we perform spatial alignment using homography and optical flow. The resulting aligned ground-truth image is well-suited for computing reference-based quantitative metrics, with the RAW capture crop serving as the input.}
\label{fig_sup:simple_align_pipeline}
\end{figure}

\section{Paired Evaluation Data Collection}
\label{sec:GT_RAW_monitor_alignment}
To use a pair of ground-truth and display capture images for reference-based evaluation of our models, the ground-truth image must first be spatially aligned with the display capture.
An overview of our alignment pipeline is depicted in \cref{fig_sup:simple_align_pipeline}.
The alignment process begins by linearizing the ground-truth image from the display domain to the sensor domain, following the method described in \cref{sec:GT_color_and_linearization}.
Since the display capture is in RAW format, we first apply bilinear demosaicing to it for the purpose of alignment.
Then, we apply the white balance gains obtained from the metadata of the RAW display capture to both the demosaiced display image and the ground-truth linearized image.

After cropping the ROI from the demosaiced display capture containing the ground-truth image, we perform bicubic downsampling on the ground-truth image such that it becomes exactly $4\times$ larger than the ROI. To correct for residual brightness difference between the two images, we match the histogram of the ground-truth image and the ROI. The calibrated homography matrix for the sensor-display is then applied to spatially aligned the ground-truth image with the RAW ROI at a global scale. 
Finally, to improve local alignment, we use a pre-trained optical flow estimation network~\cite{sun2018pwc} to further warp the ground-truth image to better match the ROI.

\begin{figure}[t]
\centering
\includegraphics[angle=0,valign=m,width=0.99\columnwidth, trim={0 0 0cm 0},clip]{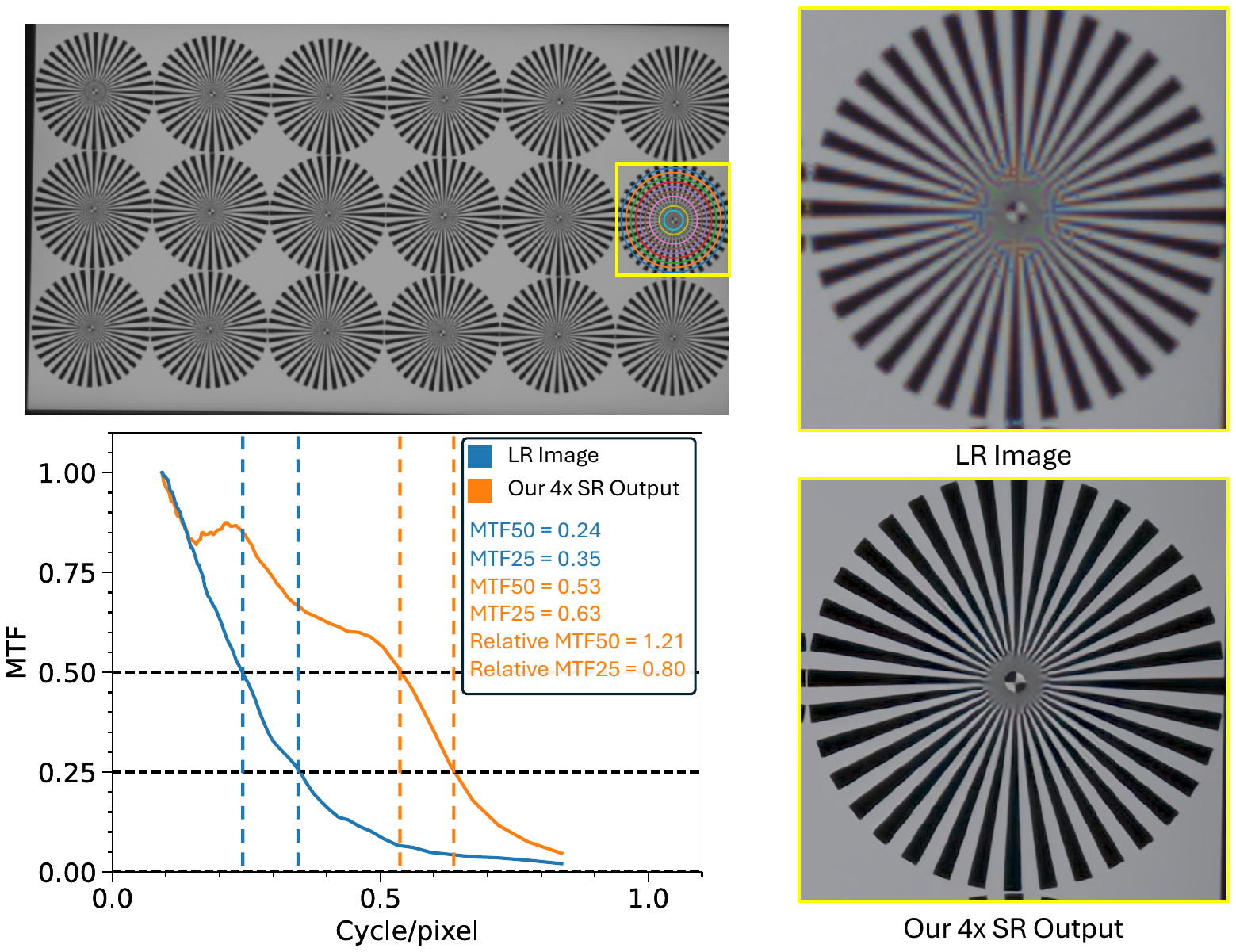} 
 \vspace{-0.3cm}
\caption{Our MTF evaluation. The MTF measured on the LR image is used to calculate the relative MTF on the output of SR models. Here is the actual measurement for our model output in the Pixel 9 Pro Main experiment (\cref{tab:per_cam_degrdation_mdel} of the main paper). The average of all values obtained over all the Siemens star patterns in the FOV is reported per model in our experiments in \cref{sec:experiments}. 
 }
\label{fig_sup:mtf_measure}
\end{figure}

\section{Details Regarding MTF Metric}
\label{secMTF_calculation}

In our experiments we use MTF50 and MTF25 to quantify the details recovered in the model outputs. 
\cref{fig_sup:mtf_measure} shows the pattern we use to measure MTF. 
The RAW capture of the displayed pattern is processed using the SR model and MTF50 and MTF25 are obtained from the MTF plot for each Siemens star in the image.
The MTF plot represents the contrast as a function of spatial frequency, derived from contrast modulations along a sinusoidal wave fitted to the intensities measured in the pixels located at different radii from the center of the pattern.
Determining the MTF curve this way can be sensitive to the distance between the camera and display. Therefore, in each camera-specific evaluation, we measure the MTF on the LR image and use it to calculate the relative MTF as a metric for the enhancement from the LR image to the super-resolved image. 
\cref{fig_sup:mtf_measure} shows MTF curves for both LR and HR images, along with the obtained relative MT50 and relative MT25 for one of the Siemens star patterns in the FOV.

\section{More Results}
\label{sub:more_for_pixel6_xiaomi}

\begin{figure*}[t]
\centering
\footnotesize
\noindent
\begin{minipage}{0.166\linewidth}
\centering
Original LR Capture 
\end{minipage}%
\begin{minipage}{0.166\linewidth}
\centering
4$\times$ Upsampling 
\end{minipage}%
\begin{minipage}{0.166\linewidth}
\centering
Real-ESRGAN~\cite{wang2021realesrgan}
\end{minipage}%
\begin{minipage}{0.166\linewidth}
\centering
BSRAW~\cite{conde2024bsraw} 
\end{minipage}%
\begin{minipage}{0.166\linewidth}
\centering
RAWSR~\cite{xu2019towards} 
\end{minipage}%
\begin{minipage}{0.166\linewidth}
\centering
Ours (Cross-camera)
\end{minipage}
\includegraphics[angle=0,valign=m,width=1\linewidth, trim={0 0 0cmx 0.8cm},clip]{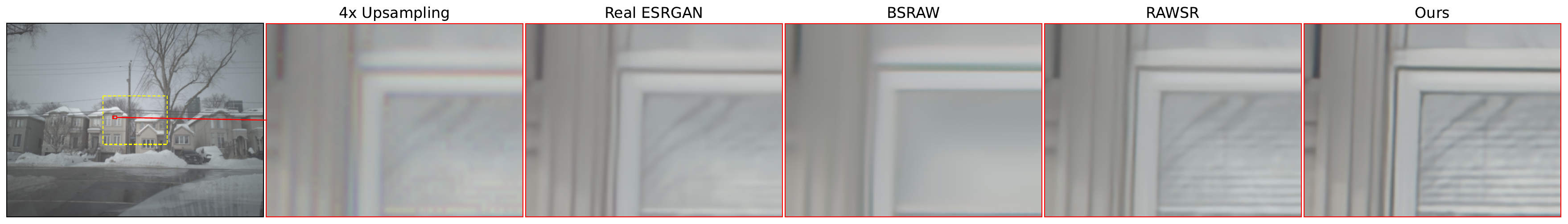} 
   \includegraphics[angle=0,valign=m,width=1\linewidth, trim={0 0 0cm 0.8cm},clip]{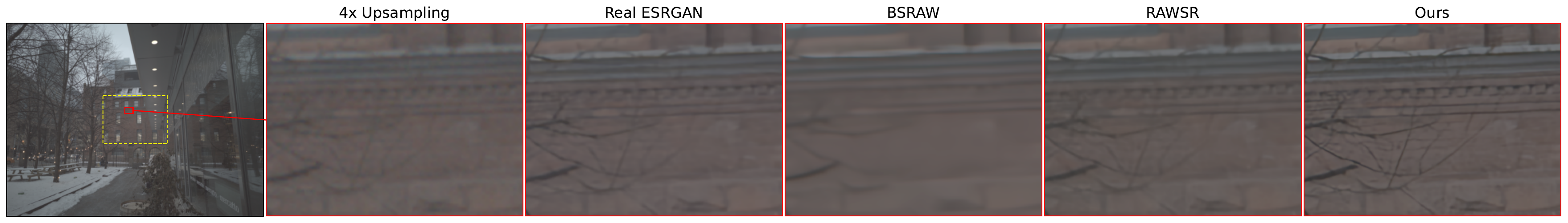} 
    \includegraphics[angle=0,valign=m,width=1\linewidth, trim={0 0 0cm 0.8cm},clip]{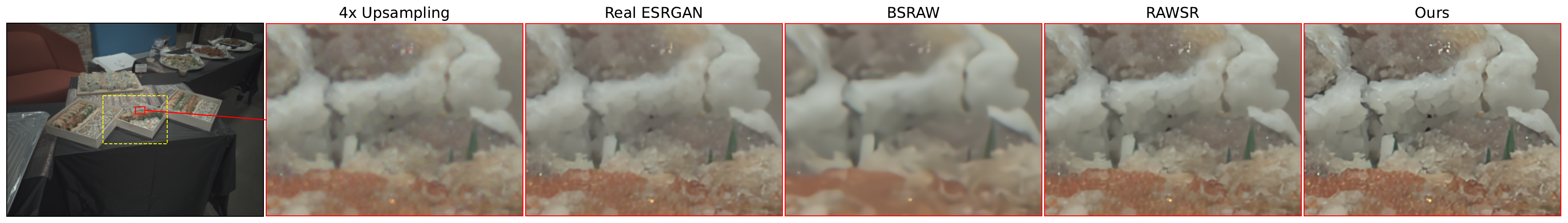}
  \includegraphics[angle=0,valign=m,width=1\linewidth, trim={0 0 0cm 0.8cm},clip]{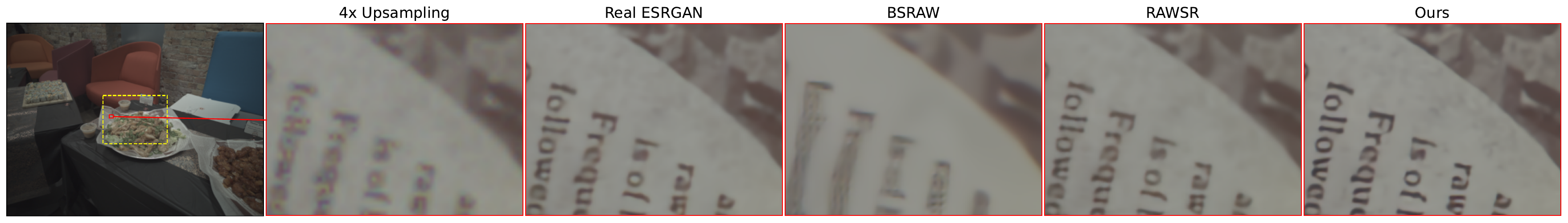}
  \includegraphics[angle=0,valign=m,width=1\linewidth, trim={0 0 0cm 0.8cm},clip]{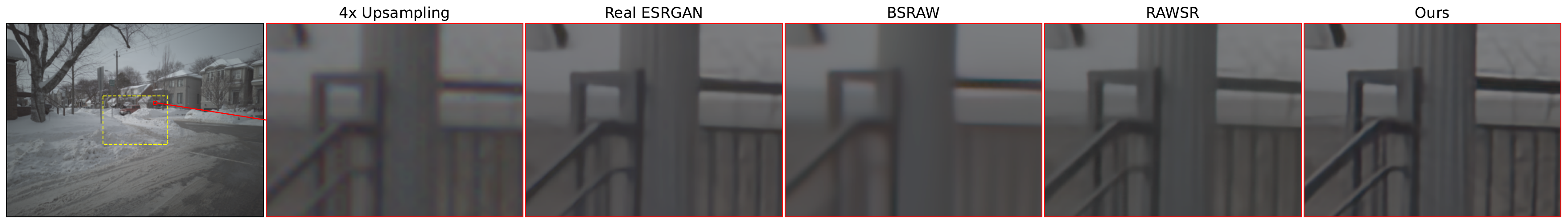} 
  \includegraphics[angle=0,valign=m,width=1\linewidth, trim={0 0 0cm 0.8cm},clip]{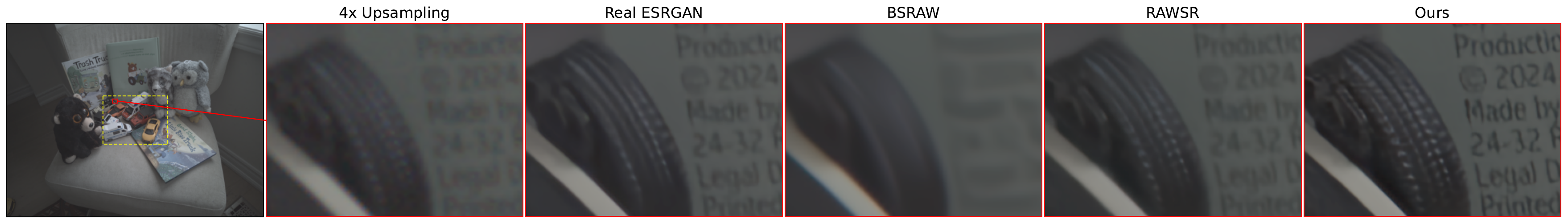}   
  \includegraphics[angle=0,valign=m,width=1\linewidth, trim={0 0 0cm 0.8cm},clip]{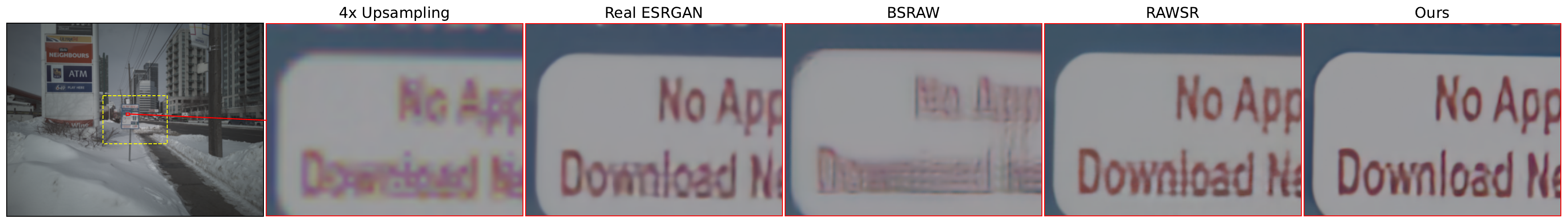}   
\includegraphics[angle=0,valign=m,width=1\linewidth, trim={0 0 0cm 0.8cm},clip]{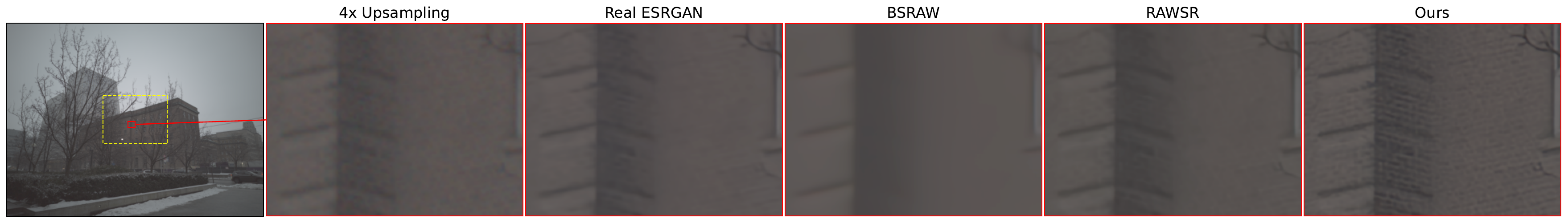}%
 \vspace{-0.3cm}
\caption{Additional qualitative results for 4$\times$ SR on \textbf{Pixel 6 Main} from the experiments in \cref{subsec:experiments_cross_camera} of the main paper. Note that Pixel 6 Main degradations are not explicitly included in the training data of any of the models. Model outputs are white-balanced and gamma-corrected better visualization.}
\label{fig_sup:more_pixel6_reults}
\end{figure*}

\begin{figure*}[t]
\centering
\footnotesize
\noindent
\begin{minipage}{0.166\linewidth}
\centering
Original LR Capture 
\end{minipage}%
\begin{minipage}{0.166\linewidth}
\centering
4$\times$ Upsampling 
\end{minipage}%
\begin{minipage}{0.166\linewidth}
\centering
Real-ESRGAN~\cite{wang2021realesrgan} 
\end{minipage}%
\begin{minipage}{0.166\linewidth}
\centering
BSRAW~\cite{conde2024bsraw}
\end{minipage}%
\begin{minipage}{0.166\linewidth}
\centering
RAWSR~\cite{xu2019towards} 
\end{minipage}%
\begin{minipage}{0.166\linewidth}
\centering
Ours (Cross-camera)
\end{minipage}
 \includegraphics[angle=0,valign=m,width=1\linewidth, trim={0 0 0cm 0.8cm},clip]{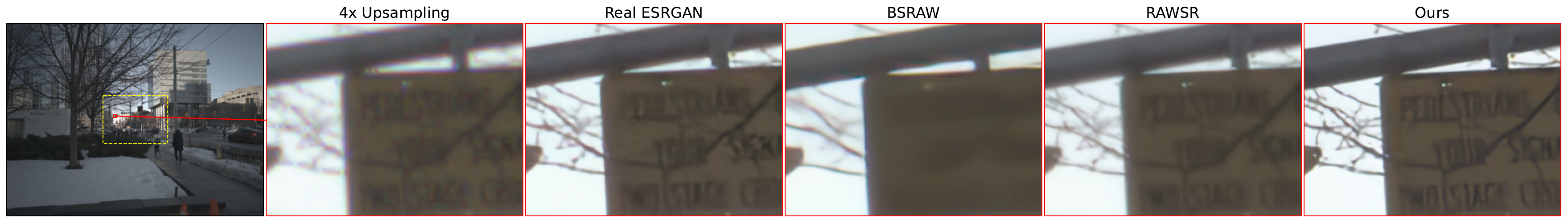} 
 \includegraphics[angle=0,valign=m,width=1\linewidth, trim={0 0 0cm 0.8cm},clip]{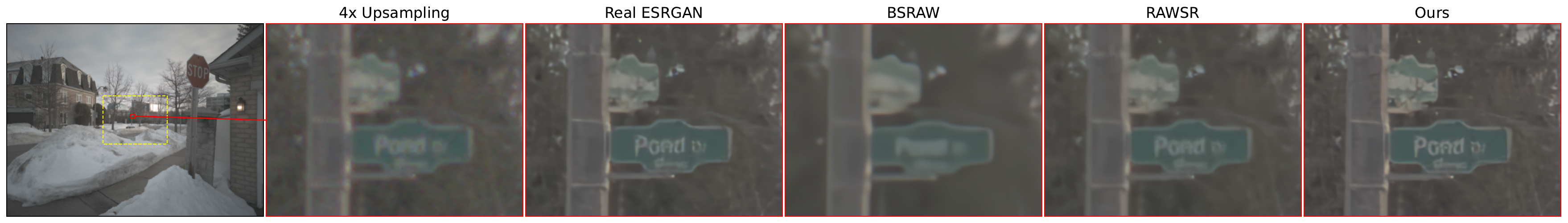} 
 \includegraphics[angle=0,valign=m,width=1\linewidth, trim={0 0 0cm 0.8cm},clip]{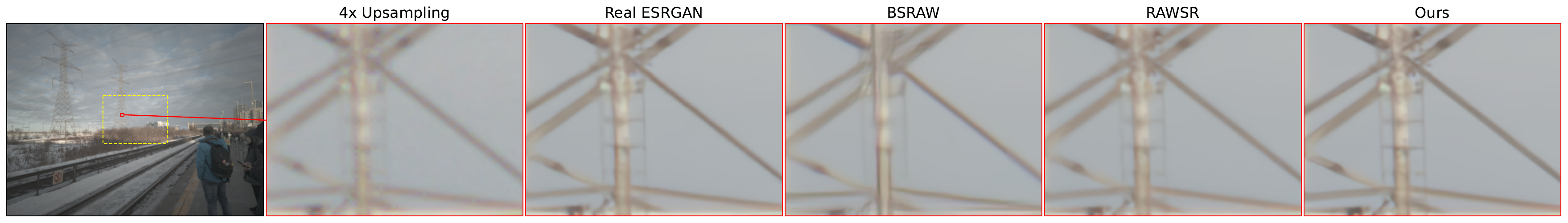} 
\includegraphics[angle=0,valign=m,width=1\linewidth, trim={0 0 0cm 0.8cm},clip]{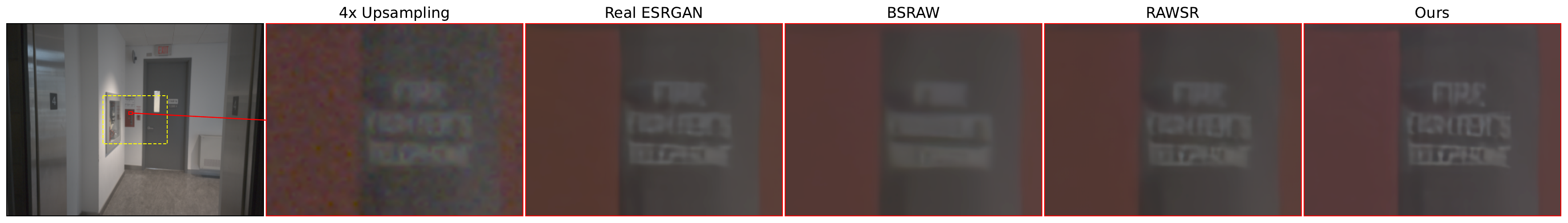} 
\includegraphics[angle=0,valign=m,width=1\linewidth, trim={0 0 0cm 0.8cm},clip]{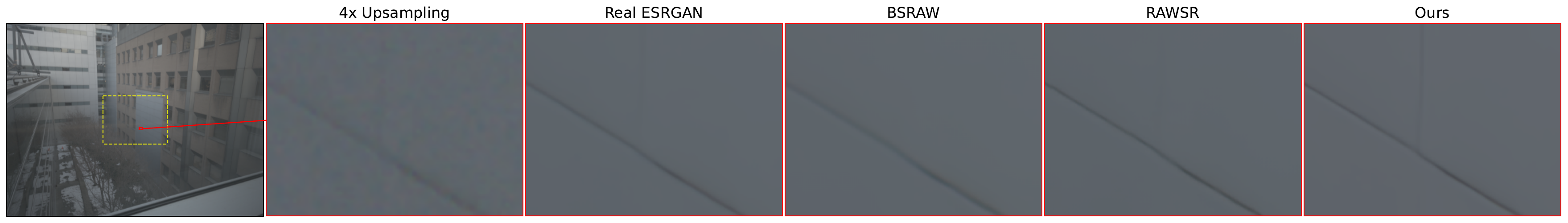} 
\includegraphics[angle=0,valign=m,width=1\linewidth, trim={0 0 0cm 0.8cm},clip]{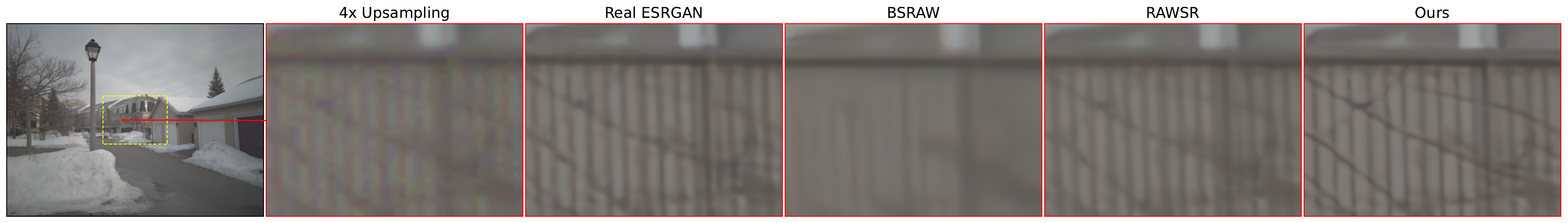} 
\includegraphics[angle=0,valign=m,width=1\linewidth, trim={0 0 0cm 0.8cm},clip]{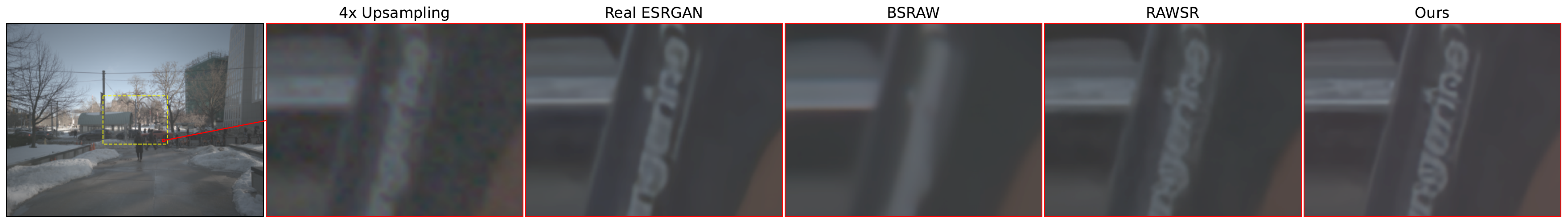}
\includegraphics[angle=0,valign=m,width=1\linewidth, trim={0 0 0cm 0.8cm},clip]{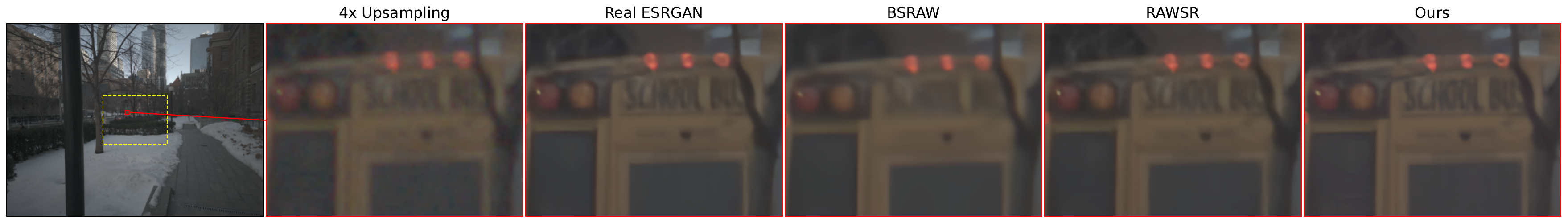}%
 \vspace{-0.3cm}
\caption{Additional qualitative results for \textbf{Mi 11 Main} 4$\times$ SR results regarding the experiments presented in Sec.~\ref{subsec:experiments_cross_camera} of the paper. Note that Mi 11 Main degradations are not explicitly seen by any of the models during training. Model outputs are white-balanced and gamma-corrected similarly for better visualization.}
\label{fig_sup:more_Xiaomi_reults}
\end{figure*}

We present additional qualitative results in \cref{fig_sup:more_pixel6_reults,fig_sup:more_Xiaomi_reults} for Pixel 6 Main and Mi 11 cameras, respectively, obtained in our cross-camera SR experiments (\cref{subsec:experiments_cross_camera} of the main paper). 

Since BSRAW~\cite{conde2024bsraw}, is one of our comparison baselines which performs RAW-to-RAW super-resolution, we convert all RGB outputs from our model into RAW format by simulating Bayer mosaicing. We then compute the PSNR and SSIM on these mosaiced outputs and compare with metrics computed on the RAW outputs from BSRAW. This approach eliminates any potential detrimental effects a downstream demosaicing method might introduce into the outputs of the BSRAW baseline. The metrics for all outputs are calculated on the 4-channel stack of Bayer channels and reported in \cref{sup_tab:blind_SR_baselines_raw_PSNR}.

\begin{table}[t]
    \centering
\small
\setlength{\aboverulesep}{0mm}
\setlength{\belowrulesep}{0mm}
\renewcommand{\arraystretch}{1.0} 
\setlength\tabcolsep{0pt}  
\begin{tabular}{
p{0.25\linewidth} 
p{0.35\linewidth} 
>{\centering}p{0.20\linewidth} 
>{\centering\arraybackslash}p{0.20\linewidth} 
}
\Xhline{2.0pt}
\multirow{2}{*}[-0.5\dimexpr \aboverulesep + \belowrulesep + \cmidrulewidth]{\makecell{Camera}}&
\multirow{2}{*}[-0.5\dimexpr \aboverulesep + \belowrulesep + \cmidrulewidth]{\makecell{Method}}& \multicolumn{2}{c}{Ref. Metrics}  \\ \cmidrule(lr){3-4}

&& PSNR            & SSIM      \\ \hline
\multirow{2}{*}[-0.5\dimexpr \aboverulesep + \belowrulesep + \cmidrulewidth]{\makecell{Pixel 6 Main}}         & BSRAW~\cite{conde2024bsraw} &  28.75               & 0.877  \\
                     & Ours (Cross-camera)       &       \textbf{29.14}    & \textbf{0.913}      \\ 
\hline
 \multirow{2}{*}[-0.5\dimexpr \aboverulesep + \belowrulesep + \cmidrulewidth]{\makecell{Mi11 Main}}         & BSRAW~\cite{conde2024bsraw}       & 33.17            &  0.897  \\
                     & Ours (Cross-camera)     & \textbf{39.72}               & \textbf{0.956}          \\                     
\Xhline{2.0pt}                     
\end{tabular}
    \caption{Our 4$\times$ SR model is trained on synthetically generated data using our degradation models obtained from seven different cameras. The performance is evaluated on RAW images captured from Pixel 6  Main and Mi 11 Main cameras, and compared with other baselines. The degradations of these two cameras are not included in the training data for any of the models.}
    \label{sup_tab:blind_SR_baselines_raw_PSNR}
\end{table}

\end{document}